\documentclass{article}
\usepackage[preprint]{preprint}
\usepackage{amssymb}
\usepackage{amsmath}
\usepackage{svg}
\usepackage{booktabs}
\usepackage{adjustbox}
\usepackage{wrapfig}
\usepackage{multicol}
\usepackage{multirow}
\usepackage{subcaption}

\title[Cell Nuclei Detection and Classification in WSIs with Transformers]{Cell Nuclei Detection and Classification in Whole Slide Images with Transformers}

\author[Pina et al.]{%
Oscar Pina\\
\institute{Universitat Politècnica de Catalunya - BarcelonaTech (UPC)}\\
\email{oscar.pina@upc.edu}\And
Eduard Dorca\\
\institute{Hospital Universitari de Bellvitge (HUB)}\\
\email{edorca@hospitalbellvitge.cat}\And
Verónica Vilaplana\\
\institute{Universitat Politècnica de Catalunya - BarcelonaTech (UPC)}\\
\email{veronica.vilaplana@upc.edu}
}

\begin{document}
\maketitle

\begin{abstract}
Accurate and efficient cell nuclei detection and classification in histopathological Whole Slide Images (WSIs) are pivotal for digital pathology applications. Traditional cell segmentation approaches, while commonly used, are computationally expensive and require extensive post-processing, limiting their practicality for high-throughput clinical settings. In this paper, we propose a paradigm shift from segmentation to detection for extracting cell information from WSIs, introducing CellNuc-DETR as a more effective solution. We evaluate the accuracy performance of CellNuc-DETR on the PanNuke dataset and conduct cross-dataset evaluations on CoNSeP and MoNuSeg to assess robustness and generalization capabilities. Our results demonstrate state-of-the-art performance in both cell nuclei detection and classification tasks. Additionally, we assess the efficiency of CellNuc-DETR on large WSIs, showing that it not only outperforms current methods in accuracy but also significantly reduces inference times. Specifically, CellNuc-DETR is twice as fast as the fastest segmentation-based method, HoVer-NeXt, while achieving substantially higher accuracy. Moreover, it surpasses CellViT in accuracy and is approximately ten times more efficient in inference speed on WSIs. These results establish CellNuc-DETR as a superior approach for cell analysis in digital pathology, combining high accuracy with computational efficiency. 
\end{abstract}

\section{Introduction}
\label{sec:intro}

Cell detection, segmentation, and classification are essential Artificial Intelligence (AI) tasks in digital pathology, enabling applications such as precise estimation of cell populations \cite{lewis2023automated}, biomarker quantification \cite{lara2021quantitative} and the spatial analysis of cells and cell graphs \cite{sobhani2021artificial, pati2022hierarchical, pina2022self, wang2024breast}. Integrating these AI-driven methods into clinical practice supports pathologists in diagnostic processes, making analyses more quantitative and reproducible \cite{garcia2016trying}. However, broader deployment of these technologies in clinical settings remains constrained by high computational demands, long processing times, and the need for robust methods capable of handling the high variability in digital pathology data.

A significant challenge in deploying AI models for digital pathology is the intensive computational resources required to analyze large-scale Whole Slide Images (WSIs). The computational burden is influenced by the model architecture—whether based on convolutional neural networks or transformer models— and the scope of the area analyzed, typically confined to patches or Regions of Interest (ROIs). Extending these analyses to entire WSIs becomes impractical in clinical settings, where rapid processing is crucial. Additionally, achieving robustness and reliability across diverse scenarios is challenging due to the variability in staining intensities, noise, and processing artifacts.

Segmentation methods, such as HoVer-Net\cite{graham2019hover}, CellViT\cite{hörst2023cellvit} and HoVer-NeXt\cite{baumann2024hover}, have recently achieved notable improvements in both accuracy and inference speed. These methods, however, use segmentation as a proxy for detection to address the difficulties in detecting small and potentially overlapping cell nuclei using traditional detection methods. Nevertheless, segmentation masks often lack clinical relevance, as the primary objective is to detect and classify cell nuclei, not to delineate their exact shapes.

To address these limitations, we propose CellNuc-DETR \cite{pina2024cell}\footnote{Originally named Cell-DETR \cite{pina2024cell}, the model has been renamed CellNuc-DETR to avoid ambiguity with a method for instance segmentation in time-lapse fluorescence microscopy images \cite{prangemeier2020c}.}, a direct detection approach for extracting cell information using the Detection Transformer (DETR) \cite{carion2020end}. Unlike traditional detection models, DETR does not rely on non-maximum suppression, allowing it to handle overlapping instances more effectively. Aligning with our hypothesis that the precise shape of cell nuclei is not essential for clinical decision-making, by focusing on detection rather than segmentation, we eliminate the computational overhead associated with generating and post-processing segmentation masks.

We demonstrate that CellNuc-DETR achieves state-of-the-art performance in both cell nuclei detection and classification across multiple datasets. Through cross-dataset evaluations on unseen datasets during training like CoNSeP \cite{graham2019hover} and MoNuSeg \cite{kumar2019multi}, we show the robustness and generalization capability of the model. Additionally, we develop an efficient inference pipeline for WSIs that significantly improves processing speed and accuracy compared to current segmentation-based approaches. This combination of speed and accuracy underscores CellNuc-DETR's potential as a practical and effective solution for clinical deployment in digital pathology. The code and the pre-trained weights are publicly available \footnote{\href{https://github.com/oscar97pina/celldetr}{https://github.com/oscar97pina/celldetr}}.

\section{Background and Related Work}
\label{sec:related}

\subsection{Transformers for Vision}
\label{sec:related:transformers}

\paragraph{Vision Transformer (ViT) \cite{dosovitskiy2020vit}} ViT is a pioneering work using transformers for vision tasks, concretely for image classification. The model tokenizes input images by partitioning them into non-overlapping patches. The patches are linearly projected and ordered in a sequence. A two-dimensional positional encoding is used to overcome the permutation invariant nature of self-attention mechanism. Additionally, a [CLASS] token is appended at the beginning of the sequence as global token for image classification after multiple transformer layers. The transformer layer architecture comprises a multi-headed self-attention module (MHA), a multi-layer perceptron (MLP) as well as two normalization layers (LN). The output of layer $(l)$ is obtained as:

\begin{align}
    \label{eq:vit}
    \mathbf{\hat{z}}^{(l)} = \text{MHA}( \text{LN}(\mathbf{z}^{(l-1)}) ) + \mathbf{z}^{(l-1)} \\
    \mathbf{z}^{(l)} = \text{MLP}( \text{LN}(\mathbf{\hat{z}}^{(l-1)}) ) + \mathbf{\hat{z}}^{(l)}
\end{align}

\paragraph{Swin Transformer (Swin) \cite{liu2021swin}} Swin is an adaptation of the transformer architecture specifically designed for images, incorporating inductive biases for tasks requiring higher resolutions such as segmentation and detection. The model introduces a hierarchical structure obtained via window attention and shifted windows (S-WMHA). Swin also partitions the image into patches, usually smaller than ViT, and performs self-attention within local windows of patches. Windows are shifted between layers to capture cross-window interactions. The model is composed of four sequential stages, each working at different resolutions and involving multiple transformer layers. At the beginning of every stage, neighboring patches are combined to create the hierarchical structure. Similarly to ViT, the output of layer $(l)$ is computed as:

\begin{align}
\label{eq:swin}
\mathbf{\hat{z}}^{(l)} = \text{S-WMHA} \left [ \text{LN} \left ( \mathbf{z^{(l-1)}} \right ) \right ] + \mathbf{z^{(l-1)}}\\
\mathbf{z}^{(l)} =  \text{MLP} \left [ \text{LN} \left ( \mathbf{\hat{z}^{(l)}} \right ) \right ] + \mathbf{\hat{z}}^{(l)}
\end{align}

\paragraph{Detection Transformer (DETR) \cite{carion2020end}} DETR introduces a novel approach to end-to-end object detection by formulating it as a direct set prediction problem. The model consists of a backbone network that extracts image features, which are then flattened into a sequence of tokens and processed through a transformer encoder. A set of learnable object queries is then fed into the transformer decoder, where they interact with the encoded features to predict bounding boxes and class labels for potential objects in the image. Each query produces a confidence score that indicates the presence of an object, allowing the model to determine which queries correspond to actual objects.

DETR is trained end-to-end using Hungarian matching \cite{kuhn1955hungarian}, a method that matches predicted objects to ground truth objects in a one-to-one manner. To ensure that all potential objects are detected, the number of object queries must exceed the maximum number of objects in an image. A key innovation of DETR is its elimination of traditional detection components such as anchors and non-maximum suppression, thereby simplifying the detection pipeline. Instead, DETR relies on the confidence scores output by each query to filter out non-object queries during inference and evaluation by applying a threshold to these scores.

\paragraph{Deformable Detection Transformer (Deformable-DETR) \cite{zhu2020deformable}} Deformable-DETR enhances DETR by replacing the global self-attention mechanism with deformable attention. Deformable attention is designed to limit the attention scope of each token to a set of learnable key sampling points around a reference point, rather than considering all possible locations. This targeted attention mechanism significantly improves the model's ability to handle small objects and densely packed scenes. Moreover, recognizing the benefits of multi-scale representations in object detection, the authors extend deformable attention to operate across multiple scales of hierarchical feature maps produced by the backbone. This extension allows Deformable-DETR to better manage objects of varying sizes and improves its overall detection performance.

\subsection{Cell Segmentation, Detection and Classification}
\label{sec:related:cell}

Given the importance of cell information for digital pathology image analysis, tasks such as cell segmentation, detection, and classification are highly relevant due to their contribution to understanding tissue architecture and disease pathology. Cell segmentation \cite{graham2019hover,baumann2024hover,anglada2024enhancing} has traditionally been more popular than detection due to the small size and frequent overlap of cell nuclei, which pose challenges for detection methods.

\paragraph{HoVer-Net \cite{graham2019hover}} HoVer-Net is a convolutional architecture designed to simultaneously solve cell segmentation and classification in histopathological images. The architecture consists of a U-Net \cite{ronneberger2015u} model with three parallel decoder branches, each addressing a different pixel-wise task. The nuclear pixel (NP) branch predicts the probability of each individual pixel belonging to a cell instance. The Horizontal-Vertical (HV) branch predicts the horizontal and vertical distances of each pixel to its corresponding cell instance center of mass, if present. The classification branch predicts the class label for each pixel. HoVer-Net effectively overcomes the overlap between cells by utilizing a post-processing step that combines the outputs of the NP and HV branches to identify distinct cell instances accurately.

While HoVer-Net demonstrates great potential, it faces several challenges in practical applications. Specifically, the model's inference time on WSIs can be considerable, which may limit its utility. Additionally, its performance on certain cell classes leaves room for improvement, suggesting that further optimizations are needed to enhance its effectiveness across a broader range of conditions.

\paragraph{CellViT \cite{hörst2023cellvit}} With the rise of transformers in vision tasks, CellViT extends HoVer-Net by incorporating a Vision Transformer (ViT) as its encoder, enhancing the model's ability to capture longer distance dependencies and improve feature representations by leveraging self-attention. Given the single-scale nature of ViTs, the model incorporates upsampling modules in the skip connections to handle multi-scale information. These skip connections are taken from different layers of the ViT. The architecture retains the three parallel decoder branches from HoVer-Net. The ViT encoder enables the model to handle different-sized images without relying on a sliding window approach for inference on larger tiles. The authors observed that, while being much more efficient, directly inferring on tiles of $1024 \times 1024$ pixels achieves similar performance to using an overlapped sliding window approach with smaller windows matching the training set image size.

The motivation of developing faster pipelines for inference on WSIs has also inspired concurrent work in the field, with different approaches being explored to address similar challenges.

\paragraph{HoVer-NeXt \cite{baumann2024hover}} HoVer-NeXt enhances the segmentation and classification performance of HoVer-Net by replacing the traditional convolutional encoder with a ConvNeXt-V2 backbone. Additionally, HoVer-NeXt incorporates test-time augmentations to further refine predictions and enhance robustness. In terms of efficiency, HoVer-NeXt significantly speeds up the inference process on WSIs. Indeed, their pipeline is $\times17$ and $\times5$ faster than HoVer-Net and CellViT, respectively. The model simplifies the three-branch decoder architecture of HoVer-Net by consolidating it into two branches: one for instance segmentation and another for classification. The instance segmentation branch is designed as a three-class pixel segmentation task, distinguishing between background, foreground, and border regions. For WSI inference, HoVer-NeXt processes larger patches compared to HoVer-Net and utilizes a "stitcher" module to combine predictions from multiple patches.

Despite the advancements introduced by HoVer-NeXt, its detection and classification performance remains lower than that of CellViT. Additionally, as segmentation-based methods, they both still require significant computational resources for post-processing, which can be particularly demanding in large WSIs. This is notable because the primary outputs of interest in digital pathology are the cell centroids and their corresponding labels, rather than the full segmentation masks.

In our work, we demonstrate that by directly performing detection rather than segmentation to extract nuclei information, it is possible to achieve superior detection and classification performance while also significantly improving computational efficiency. Detection methods tend to be less computationally intensive than segmentation, and the absence of post-processing further streamlines the pipeline. This increased efficiency and accuracy make our approach particularly well-suited for implementation in clinical practice, where quick and reliable analysis of WSIs is crucial for diagnostic and prognostic decision-making.
\section{Materials and Methods}
\label{sec:methods}

\subsection{Datasets}
\label{sec:methods:datasets}

In this section, we describe the datasets used in our experiments. Each dataset serves a specific purpose in evaluating and enhancing models for cell nuclei detection, classification, and inference efficiency.

\paragraph{TCGA} The Cancer Genome Atlas (TCGA) is a public repository that contains a large collection of Whole Slide Images (WSIs) from a wide variety of cancer types. The dataset includes histopathological images from over 11,000 patients, representing diverse tumor types such as lung, breast, colon, prostate, and kidney cancers. To evaluate the efficiency of the CellNuc-DETR pipeline and compare it with existing cell segmentation methods in real-world scenarios, we utilize 20 WSIs from TCGA and report the inference times. 

\begin{figure}
    \centering
    \includegraphics[width=0.45\textwidth]{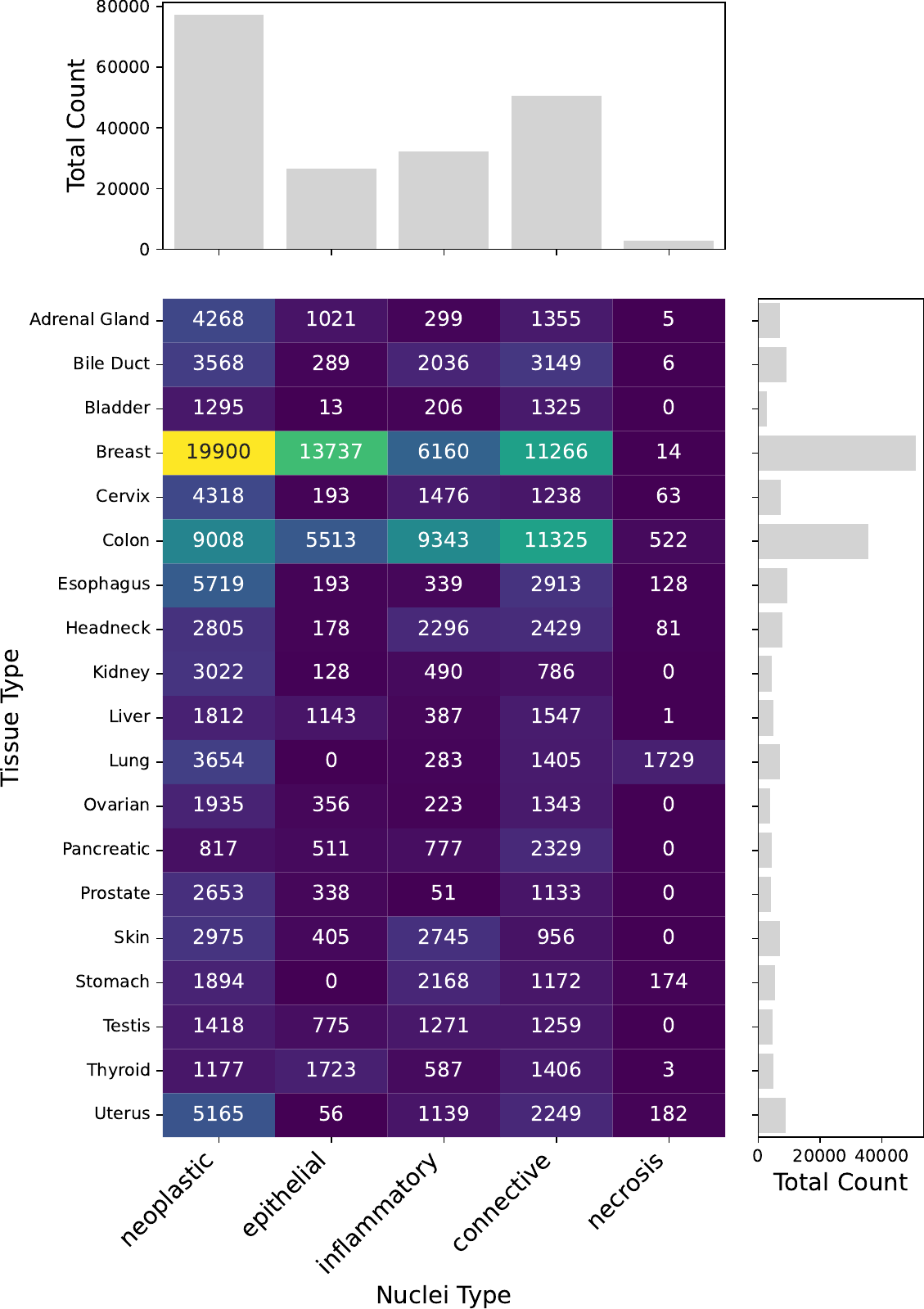}
    \caption{Cell nuclei class distribution across tissues on the PanNuke dataset.}
    \label{fig:pannuke-stats}
\end{figure}

\paragraph{PanNuke} 
The PanNuke dataset \cite{gamper2020pannuke} consists of 7,904 image patches, each with dimensions of $256 \times 256$ pixels, derived from WSIs within TCGA. These patches cover 19 different tissue types, all captured at a $40\times$ magnification. The dataset includes detailed annotations for 189,744 nuclei, classified into five clinically relevant categories: neoplastic, inflammatory, connective, necrotic, and epithelial. Figure \ref{fig:pannuke-stats} illustrates the distribution of cell counts across tissues and nucleus types, highlighting the significant class imbalance both within nucleus types and across different tissues. This variability in cell count and distribution makes PanNuke a challenging dataset, but also a valuable resource for addressing generalization in digital pathology. Such variability is crucial for developing algorithms that can perform well across different tissue types and staining protocols. The dataset is divided into three predefined folds for fair model comparison, with fold partitioning designed to ensure each fold contains an equal portion of the smallest class \cite{gamper2020pannuke}.


\paragraph{CoNSeP}
The CoNSeP dataset \cite{graham2019hover} comprises 41 tiles, each measuring $1000 \times 1000$ pixels, sourced from H\&E-stained colorectal adenocarcinoma WSIs at a $40 \times$ magnification. This dataset is notably diverse, covering various tissue regions including stromal, glandular, muscular, collagenous, adipose, and tumorous areas. It features a wide array of nuclei from different cell types, initially labeled as normal epithelial, malignant/dysplastic epithelial, fibroblast, muscle, inflammatory, endothelial, or miscellaneous. The miscellaneous category includes necrotic, mitotic, and unclassifiable cells. Following previous work \cite{graham2019hover}, we consolidated the normal and malignant/dysplastic epithelial nuclei into a single class (epithelial), and combined the fibroblast, muscle, and endothelial nuclei into a class named spindle-shaped nuclei. The dataset is divided into predefined training and validation sets to facilitate model development and evaluation.

\paragraph{MoNuSeg}
The MoNuSeg dataset \cite{kumar2019multi} includes 44 images, each with dimensions of 1000 × 1000 pixels at a $40 \times$ magnification. Sourced from H\&E-stained tissue samples from various organs in the TCGA archive, this dataset provides detailed annotations of nuclear boundaries. It is divided into a training set with 30 images, containing around 22,000 nuclear boundary annotations, and a test set with 14 images, including an additional 7,000 annotations. Notably, this dataset does not provide class labels for the nuclei, focusing solely on detection and segmentation.


The datasets used in our study provide image patches of varying sizes, each tailored to specific research objectives. The PanNuke dataset consists of patches sized $256 \times 256$ pixels at 0.25 $\mu$m/px, translating to an area of 4096 $\mu$m$^2$. These patches are employed for both training and evaluating cell nuclei detection and classification models. The CoNSeP and MoNuSeg datasets offer larger patches, each measuring 1000$\times$1000 pixels, covering an area of 62,500 $\mu$m$^2$. These datasets are utilized for cross-domain evaluation to assess the generalizability of our models. The slides extracted from the TCGA have an average area of 300 mm$^2$ per slide, used specifically to evaluate the inference times of the models. This approach ensures robust training, evaluation, and performance assessment across diverse tissue types and scales.

\subsection{Cell Nuclei detection and Classification with Transformers}
\label{sec:methods:CellNuc-DETR}

\subsubsection{Architecture}
\label{sec:methods:CellNuc-DETR:architecture}

\begin{figure}
    \centering
    \includegraphics[width=0.45\textwidth]{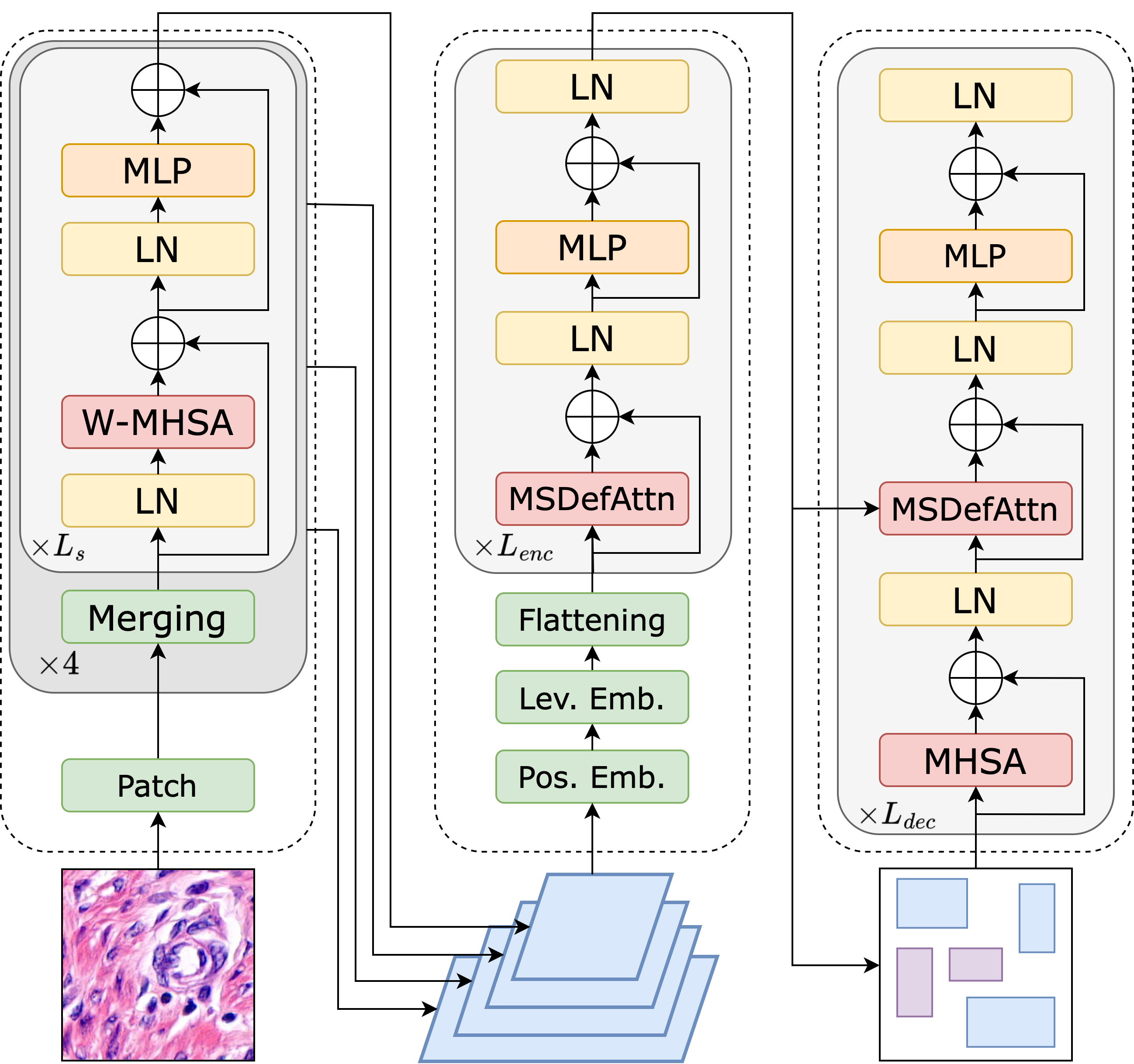}
    \caption{CellNuc-DETR model architecture.}
    \label{fig:architecture}
    \footnotesize{The model consists of a hierarchical backbone, such as a Swin Transformer, which outputs feature maps at multiple resolutions. These maps are flattened and embedded with positional and level information to form input sequences for the deformable transformer encoder. The encoder processes these sequences using multi-scale deformable attention (MSDefAttn). It then generates initial bounding boxes, which are used as input queries for the decoder and as memory for the deformable cross-attention modules. The model outputs bounding box and label predictions for each input image.}
\end{figure}

The architecture employed in our study adopts a hierarchical backbone that generates a multi-level feature pyramid from input images. This is followed by a multi-scale deformable transformer \cite{zhu2020deformable}, including the encoder and decoder components. Figure \ref{fig:architecture} shows the model architecture. The encoder enhances input features through multi-scale deformable self-attention mechanisms, while the decoder outputs predictions for bounding boxes and labels based on a set of object queries. These queries are initialized to represent potential object locations within the input images following the two-stage approach  \cite{zhu2020deformable}. Both the backbone and transformer components are pretrained on large-scale datasets to capture diverse feature representations. Our experiments explore various backbone architectures and transformer configurations to assess their impact on detection and classification performance.

\subsubsection{Training}
\label{sec:methods:CellNuc-DETR:training}

\paragraph{Model initialization}
The backbone and the deformable transformer utilized in our experiments are pretrained on ImageNet \cite{deng2009imagenet} and Common Objects in Context (COCO) \cite{lin2014microsoft} datasets, respectively. However, the neck connecting the backbone and the transformer, as well as the classification layers of the transformer, are initialized using a random strategy.

\paragraph{Data augmentation}
Digital pathology images exhibit substantial diversity due to various factors, including differences in staining protocols, elapsed time since slide staining before digitization, and the diverse tissue types. This variability poses challenges for model generalization and performance in tasks such as cell nuclei detection and classification. Data augmentation serves as strategy to mitigate these challenges by enriching the dataset with diverse representations. Drawing inspiration from insights in \cite{10.1117/12.2293048}, our augmentation pipeline combines traditional techniques such as rotation, flipping, color jittering, and blurring with advanced stain augmentations. Specifically, we transform RGB images into Hematoxylin-Eosin-DAB (HED) space, independently manipulate channels to simulate staining variations, and then revert to RGB format. This approach ensures our models are trained on a robust dataset that captures the complexities and variability inherent in digital pathology images, thereby enhancing their ability to generalize and perform effectively across different conditions.

\paragraph{Loss function}
In our study, we utilize the standard loss function recommended for Detection Transformers in natural images, which includes three components: bounding box L1 regression loss, generalized intersection over union (GIoU) loss, and focal loss for cell nuclei classification.

The bounding box L1 regression loss measures the difference between predicted and ground truth bounding box coordinates, ensuring precise localization of cell nuclei. To address scale dependency, GIoU loss is incorporated. Finally, given the imbalanced nature of cell nuclei classification, focal loss is used to handle this challenge by down-weighting well-classified examples and focusing on hard-to-classify instances. This enhances the model’s ability to accurately classify diverse cell types.

\paragraph{Optimization}
Training is distributed across four NVIDIA Quadro RTX 16GB GPUs to expedite computation. We employ the Adam optimizer with a base learning rate of $2 \times 10^{-4}$ defined for a batch size of 16 in the original paper, which we linearly scale based on our setting. The learning rate for the parameters in the backbone and the multi-scale deformable attention modules are initialized at $2 \times 10^{-5}$. A weight decay of $1 \times 10^{-4}$ is applied to prevent overfitting. Hyperparameter adjustment accommodates our specific batch size and GPU setup, aligning with practices established in related works. Learning rate scheduling follows a multi-step approach, reducing the base learning rate by a factor of 0.1 at 70\% and 90\% of the total training duration to stabilize convergence and enhance model performance. All models are trained for 100 epochs to ensure comprehensive convergence and evaluation across datasets and experimental conditions.

\paragraph{Hyperparameters}
We adopt configurations based on the original Deformable DETR framework \cite{zhu2020deformable}, leveraging established hyperparameters to avoid exhaustive search. These configurations include optimization hyperparameters such as learning rate and weight decay, as well as loss weights for the composite loss function. Minor adjustments were made to align with our specific batch size and GPU setup. Additionally, we modified the learning rate scheduler based on related work to better suit our digital pathology tasks.

\subsubsection{Inference pipeline}
\label{sec:methods:CellNuc-DETR:inference}

\begin{figure}[h!]
    \centering
    \begin{subfigure}[t]{\textwidth}
        \centering
        \includegraphics[width=\textwidth]{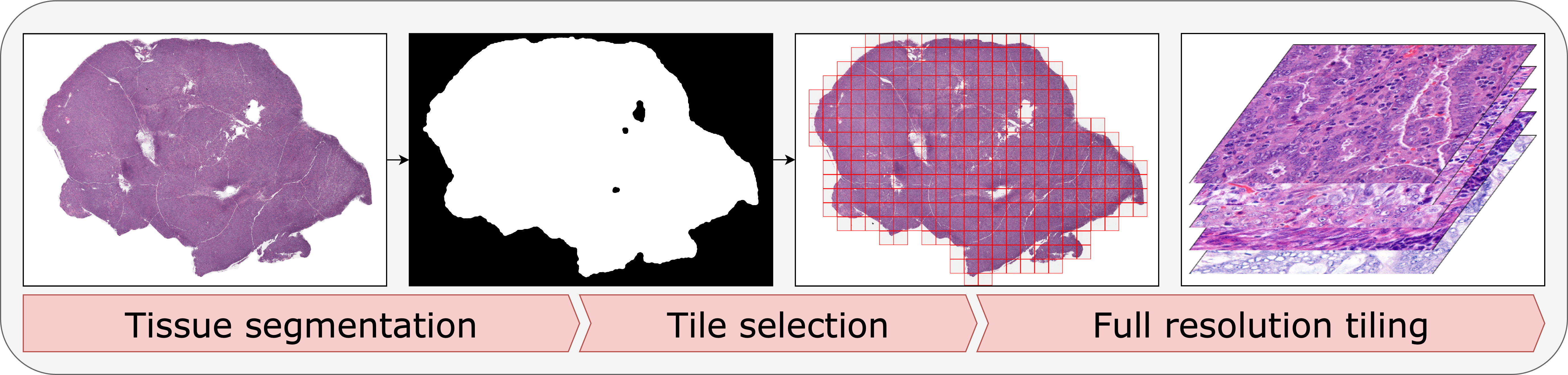} 
    \end{subfigure}
    

    \begin{subfigure}[t]{\textwidth}
        \centering
        \includegraphics[width=\textwidth]{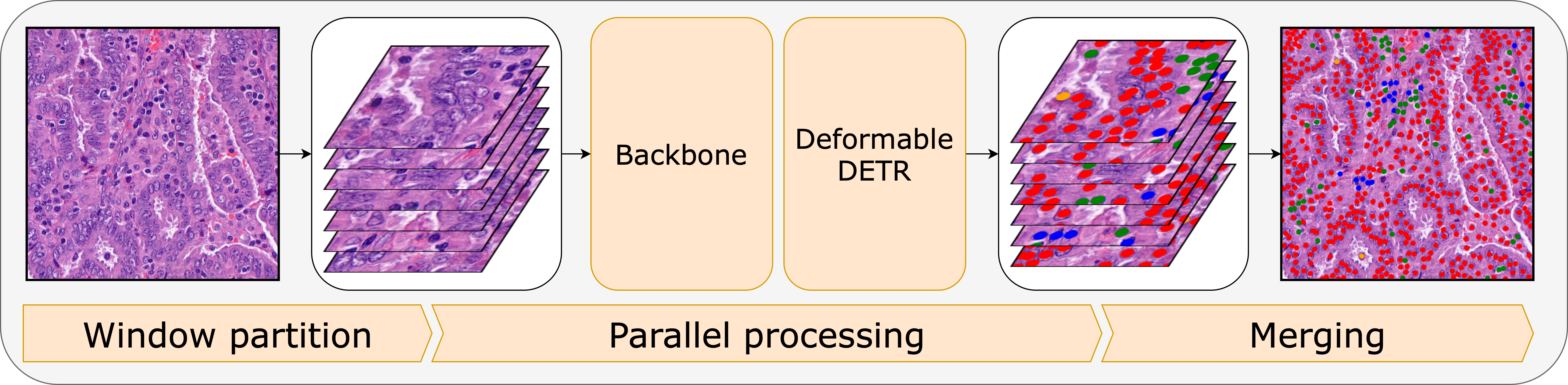} 
    \end{subfigure}
    

    \begin{subfigure}[t]{\textwidth}
        \centering
        \includegraphics[width=\textwidth]{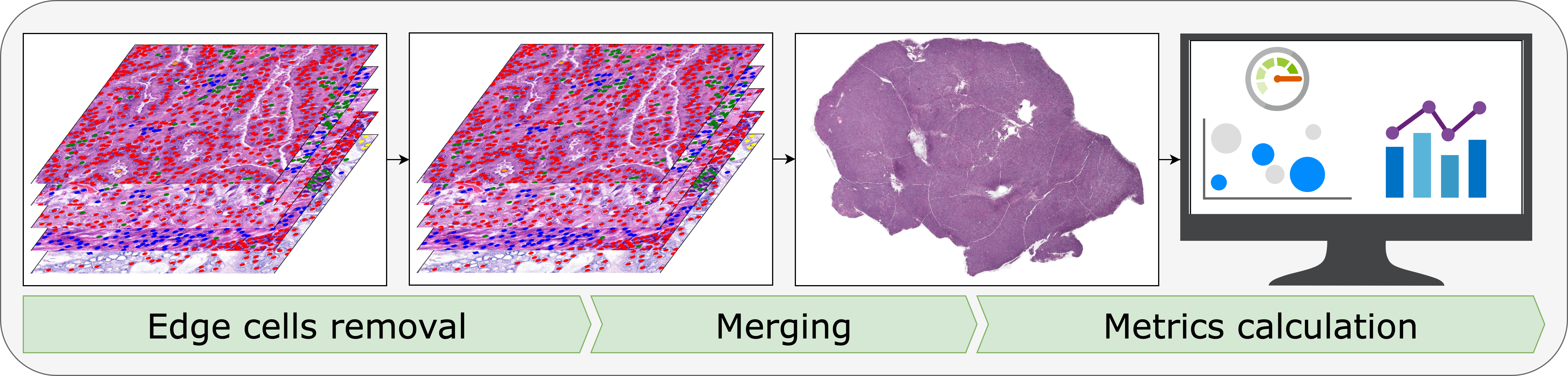} 
    \end{subfigure}
    
    \caption{CellNuc-DETR inference pipeline on WSIs.}
    \label{fig:pipeline}
    \footnotesize{
    \textit{(1) Pre-processing:}  Tissue segmentation removes background regions, and full-resolution tiles are extracted. \textit{(2) Inference:} Each tile is partitioned into overlapping windows and processed in parallel by the model, then predictions are combined. \textit{(3) Post-processing:} Predictions from all tiles are combined, with edge cells outside central borders removed to avoid duplication.
    }
\end{figure}

The inference pipeline for CellNuc-DETR on Whole Slide Images (WSIs) involves three main steps: pre-processing, inference, and post-processing, as illustrated in Figure \ref{fig:pipeline}. In this section, we detail each of these steps, highlighting how they contribute to efficient and accurate cell nuclei detection.

\paragraph{Pre-processing} 
The giga-size of WSIs limits the feasibility of end-to-end processing pipelines, necessitating the subdivision of these images into smaller tiles for independent analysis. Additionally, many areas within WSIs are non-informative background regions, making it essential to detect and ignore these sections to optimize computational resources.

To address these challenges, we first detect the tissue regions within the WSIs. We convert the WSI thumbnail into the HED color space \cite{ruifrok2001quantification}, enhancing the contrast between tissue and background. By applying specific thresholds to each channel in the HED space, we create a binary mask that isolates the tissue regions from the non-tissue areas. After identifying the tissue regions, we subdivide the WSIs into overlapping tiles. The overlap ensures comprehensive coverage and captures sufficient contextual information, which is crucial for areas with high cell density. This approach also facilitates seamless merging of predictions in later stages.

The pre-processing step returns the top-left corner coordinates of the tiles to be processed, guiding the subsequent stages and ensuring that all relevant regions of the WSI are analyzed efficiently and accurately.

\paragraph{Large tile inference} A significant constraint of DETR-like models is the necessity for the number of queries in the decoder to surpass the potential objects present in an image. In regions characterized by a high cell density, a $256 \times 256$px image patch may contain up to 300 cell nuclei. Consequently, increasing the input image size to larger tiles, such as $1024 \times 1024$px, becomes non-trivial. The number of cell nuclei, and therefore the required input DETR queries, can substantially increase, potentially resulting in prohibitive computational demands.

\begin{figure}
    \centering
    \includegraphics[width=0.5\textwidth]{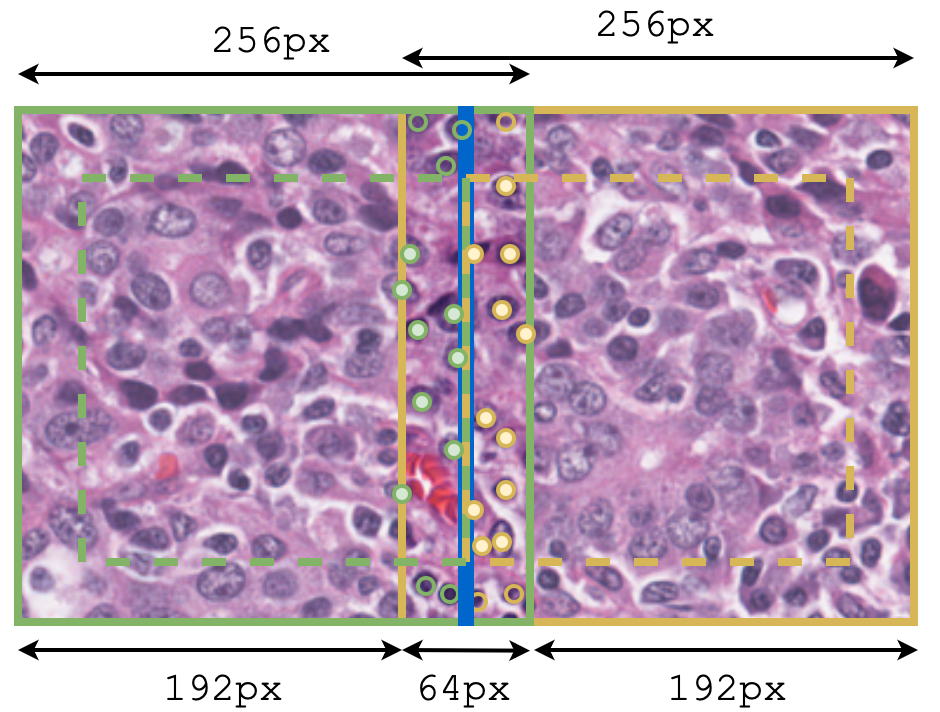}
    \caption{Resolution of edge cells between overlapped windows.}
    \footnotesize{
        The plotted cells are detected during the processing of both the green and the yellow windows. However, only those detections inside the central crop of the windows are considered. The filled green (yellow) centroids will be assigned to the left (right) window, whereas the unfilled ones will only be considered if there are no windows above or below.
    }
    \label{fig:edge-cells}
\end{figure}

To address this challenge during inference on larger image tiles, we adopt a simultaneous processing of overlapped sliding windows approach. The model, trained on smaller patches (e.g., $256 \times 256$px), processes larger images by dividing them into overlapping windows. These windows are processed in-device, minimizing GPU-CPU communication and enhancing inference speed. Specifically, the model splits the original image into overlapped windows, processes them in parallel, and then combines the outputs to derive the final results. To merge predictions from overlapping windows, we only keep the centroids within the central crop of each window, leaving a border of half the overlap size on each side. This strategy ensures that detections near the window borders are excluded to avoid redundancy, as they are covered by the central regions of adjacent windows. For windows at the edges of the tiles, this exclusion applies solely to the sides overlapped by another window, not to those corresponding to the image borders. The process is illustrated in Figure \ref{fig:edge-cells}. This approach is faster than pre-processing image patches before sending them to the device, and its efficiency is particularly beneficial when processing WSIs. By employing this strategy, CellNuc-DETR effectively manages the computational demands of high-density cell regions and large images, ensuring robust performance in practical applications.

\paragraph{Slide inference} The slides, which consist of multiple overlapping tiles, are processed using this approach. Tiles are dynamically retrieved from the slide using OpenSlide \cite{goode2013openslide} based on the coordinates obtained in the pre-processing step. By retrieving large tiles from the WSI and partitioning them into overlapping windows directly on the device, we achieve a more efficient workflow. This method is faster than retrieving small patches from the slide or partitioning large tiles into overlapping windows before sending them to the device. The efficiency gains come from minimizing interactions with the slide (on disk) and reducing CPU-GPU communication, resulting in a optimized inference process.

\paragraph{Post-processing}
Given that the tiles are also defined with overlap, we follow a strategy similar to the sliding window approach to combine predictions between tiles. After processing the tiles, we merge the predictions by only keeping the centroids within the central crop of each tile, excluding detections near the borders to avoid redundancy. This method leverages the overlap to ensure consistent and accurate merging of detections across tile boundaries. As the post-processing for cell nuclei detection is relatively simple, involving just the merging of centroid coordinates, it is much faster compared to the complex post-processing required for segmentation methods. This efficiency in post-processing significantly enhances the overall performance, making it well-suited for practical applications in digital pathology.

By default, we define a tile size of $1024 \times 1024$px with an overlap of 64px. Each tile is divided into windows of size $256 \times 256$px with an overlap of 64px. As the overlap between tiles is the same as the overlap between windows, the result after merging the predictions is agnostic to the tile partitioning.

\section{Experiments and Results}
\label{sec:results}

\subsection{Experimental Design}
\label{sec:results:design}

\subsubsection{Experiments}
\label{sec:results:design:experiments}

The experiments are designed to evaluate the performance, generalization, robustness, and efficiency of CellNuc-DETR in various contexts. First, we train CellNuc-DETR using the three-fold cross-validation split provided in the PanNuke dataset, comparing its performance against other state-of-the-art models in terms of cell nuclei detection and classification metrics. During this stage, we also assess different design components of the model, such as the input and backbone feature resolution, the model depth, and other architectural variations. 

Next, we train CellNuc-DETR on 80\% of the PanNuke dataset, reserving the remaining 20\% for validation. This model is used in subsequent experiments to evaluate its performance across different datasets and scenarios.

To test the robustness of CellNuc-DETR, we perform cross-dataset evaluations on the MoNuSeg and CoNSeP datasets using the pre-trained model without additional fine-tuning. For the MoNuSeg dataset, which lacks classification labels, we evaluate detection performance only. For the CoNSeP dataset, we map the five classes predicted by PanNuke to the four classes present in CoNSeP to evaluate both detection and classification performance. These evaluations also assess the effectiveness and adaptability of the sliding window inference approach, as these datasets comprise larger image tiles.

Finally, we perform inference on the TCGA slides to measure the inference time performance of CellNuc-DETR. This experiment focuses on the efficiency and scalability of the model when applied to WSIs, highlighting its practicality for real-world applications.

\subsubsection{Evaluation Metrics}
\label{sec:results:design:metrics}

The evaluation protocol for nuclei detection and classification follows the methodology outlined in \cite{graham2019hover}, employing F1-score as the evaluation metric. Initially, a bi-partite matching process aligns ground truth nuclei centroids with detected counterparts, limited to a radius of 12 pixels for a resolution of 0.25 $\mu$m/px. Detection metrics, including true positives ($TP_{det}$), false positives ($FP_{det}$), and false negatives ($FN_{det}$), are derived based on the outcomes of the matching process between ground truth and predicted nuclei. The detection F1-score ($F_{det}$) is computed as the harmonic mean of detection precision ($P_{det}$) and recall ($R_{det}$).

For classification, $TP_{det}$ is further categorized into correctly and incorrectly classified nuclei of class \textit{c}, denoted as $TP_c$ and $FP_c$, respectively. Additionally, misclassified elements from class $c$ are captured as $FN_c$. Precision, Recall, and F1-Score for each class are then calculated as follows:
\begin{equation}
    F_c = \frac{2(TP_c +TN_c)}{2(TP_c +TN_c)+2FP_c +2FN_c +FP_{det} +FN_{det}}
\end{equation}

\begin{equation}
    P_c = \frac{TP_c +TN_c}{TP_c +TN_c +2FP_c +FP_{det}}
\end{equation}

\begin{equation}
    R_c = \frac{TP_c +TN_c}{TP_c +TN_c +2FN_c +FN_{det}}
\end{equation}

\subsection{Numerical Results}
\label{sec:results:numerical}

\subsubsection{Evaluation on PanNuke}
\label{sec:results:numerical:pannuke}

In Table \ref{tab:pannuke}, we evaluate CellNuc-DETR and compare it with other state-of-the-art methods. Our results demonstrate that CellNuc-DETR outperforms these methods in both cell nuclei detection and classification, achieving state-of-the-art performance.

We conducted an evaluation of different design components in our study. Specifically, we evaluated the performance of our models with four different backbones: ResNet-50 \cite{he2016deep} (\textit{R50}), Swin-Tiny (\textit{tiny}), Swin-Base (\textit{base}), and Swin-Large (\textit{large}) \cite{liu2021swin}. Additionally, given the small size and potential overlap between cell nuclei, we assessed the influence of the resolution of the feature maps taken from the backbone, comparing the usage of all four available levels (\textit{4lvl}) to using only the last three levels (\textit{3lvl}), as done in the original Deformable DETR \cite{zhu2020deformable}. Finally, we investigated the complexity of the model by varying the number of decoder queries and layers. The original DETR, trained on the COCO dataset, uses 300 decoder queries and selects the top 100 predictions, as COCO images contain a maximum of 100 objects. For cell nuclei detection, where there may be up to 300 cells in an image, we used 900 queries and selected the top 300. This model is composed by six encoder and decoder layers (Deep). We also experimented with using 600 queries and selecting the top 200, as it is rare to have 300 cells in a single image. The model using 600 queries also employed only three encoder and decoder layers, rather than six as in the original architecture.

The results show that the backbone is crucial for achieving state-of-the-art performance in both cell nuclei detection and classification tasks. Specifically, models using the Swin backbone consistently outperform those with the ResNet-50 backbone across all classes. Among the Swin-based models, Swin-Large shows a slight better performance. The number of feature levels extracted from the backbone did not significantly influence the results, which contrasts with findings from previous studies. Additionally, the deepest models with 900 queries achieve state-of-the-art performance in cell classification, slightly surpassing the smaller models. However, the marginal difference suggests that the smaller models could be favored for faster inference without a loss in accuracy, offering a more efficient alternative.


\begin{table}[ht]
\vskip -0.1in
\centering
\caption{Detection and classification metrics on PanNuke dataset.}
\label{tab:pannuke}
\vskip 0.1in
\begin{adjustbox}{width=\textwidth}
\begin{tabular}{l|ccc|ccc|ccc|ccc|ccc|cccccc}
\toprule
\multirow{2}{*}{\textbf{Method}} & \multicolumn{3}{c|}{\textbf{Detection}} & \multicolumn{3}{c|}{\textbf{Neoplastic}} & \multicolumn{3}{c|}{\textbf{Epithelial}} & \multicolumn{3}{c|}{\textbf{Inflammatory}} & \multicolumn{3}{c|}{\textbf{Connective}} & \multicolumn{3}{c}{\textbf{Necrosis}} \\
& $P_{det}$ & $R_{det}$ & $F_{det}$ & $P_{neo}$ & $R_{neo}$ & $F_{neo}$ & $P_{epi}$ & $R_{epi}$ & $F_{epi}$ & $P_{inf}$ & $R_{inf}$ & $F_{inf}$ & $P_{con}$ & $R_{con}$ & $F_{con}$ & $P_{nec}$ & $R_{nec}$ & $F_{nec}$ \\
\midrule
\textbf{DIST} \cite{naylor2018segmentation} & 0.74 & 0.71 &0.73 & 
       0.49 & 0.55 & 0.50 & 
       0.38 & 0.33 & 0.35 &
       0.42 & 0.45 & 0.42 &
       0.42 & 0.37 & 0.39 &
       0.00 & 0.00 & 0.00 \\
\textbf{Mask-RCNN} \cite{he2017mask} & 0.76 & 0.68 & 0.72 &
            0.55 & 0.63 & 0.59 &
            0.52 & 0.52 & 0.52 &
            0.46 & 0.54 & 0.50 &
            0.42 & 0.43 & 0.42 &
            0.17 & 0.30 & 0.22 \\
\textbf{Micro-Net} \cite{raza2019micro} & 0.78 & 0.82 & 0.80 &
            0.59 & 0.66 & 0.62 &
            0.63 & 0.54 & 0.58 &
            0.59 & 0.46 & 0.52 &
            0.40 & 0.45 & 0.47 &
            0.23 & 0.17 & 0.19 \\
\textbf{HoVer-Net} \cite{graham2019hover} & 0.82 & 0.79 & 0.80 &
            0.58 & 0.67 & 0.62 &
            0.54 & 0.60 & 0.56 &
            0.56 & 0.51 & 0.54 &
            0.52 & 0.47 & 0.49 &
            0.28 & 0.35 & 0.31 \\

\textbf{HoVer-NeXt}$_{tiny, TTA=4}$ \cite{baumann2024hover} & 0.82 & 0.77 & 0.79 &
        0.68 & 0.61 & 0.65 &
        0.67 & 0.64 & 0.65 &
        0.55 & 0.57 & 0.56 &
        0.53 & 0.50 & 0.51 &
        0.41 & 0.34 & 0.37 \\

\textbf{CellViT}$_{256}$ \cite{hörst2023cellvit} & 0.83 & 0.82 & 0.82 &
          0.69 & 0.70 & 0.69 &
          0.68 & 0.71 & 0.70 &
          0.59 & 0.58 & 0.58 &
          0.53 & 0.51 & 0.52 &
          0.39 & 0.35 & 0.37\\

\textbf{CellViT}$_{SAM-H}$ \cite{hörst2023cellvit} & 0.84 & 0.81 & 0.83 &
          0.72 & 0.69 & 0.71 &
          0.72 & 0.73 & 0.73 &
          0.59 & 0.57 & 0.58 &
          0.55 & 0.52 & 0.53 &
          0.43 & 0.32 & 0.36\\

\midrule

\textbf{CellNuc-DETR}$_{R50,3lvl}$-Deep&0.81&0.85&0.83&0.67&0.71&0.69&0.67&0.72&0.69&0.56&0.61&0.58&0.52&0.53&0.52&0.45&0.37&0.40\\
\textbf{CellNuc-DETR}$_{R50,4lvl}$-Deep&0.81&0.85&0.83&0.67&0.71&0.69&0.68&0.71&0.70&0.56&0.62&0.59&0.53&0.53&0.53&0.47&0.39&0.43\\
\midrule
\textbf{CellNuc-DETR}$_{tiny,3lvl}$&0.81&0.85&0.83&0.69&0.73&0.71&0.68&0.75&0.71&0.58&0.62&0.60&0.55&0.55&0.55&0.48&0.42&0.44\\
\textbf{CellNuc-DETR}$_{tiny,3lvl}$-Deep&0.81&0.87&0.84&0.69&0.74&0.71&0.68&0.77&0.72&0.57&0.64&0.60&0.54&0.56&0.55&0.50&0.43&0.46\\
\textbf{CellNuc-DETR}$_{tiny,4lvl}$&0.80&0.86&0.83&0.68&0.73&0.71&0.67&0.76&0.71&0.57&0.64&0.60&0.55&0.54&0.55&0.45&0.43&0.44\\
\textbf{CellNuc-DETR}$_{tiny,4lvl}$-Deep&0.80&0.87&0.84&0.69&0.74&0.71&0.68&0.77&0.72&0.56&0.65&0.61&0.55&0.55&0.55&0.49&0.44&0.46\\

\midrule

\textbf{CellNuc-DETR}$_{base,3lvl}$&0.81&0.86&0.83&0.68&0.73&0.71&0.69&0.75&0.72&0.57&0.63&0.60&0.54&0.55&0.55&0.48&0.41&0.44\\
\textbf{CellNuc-DETR}$_{base,3lvl}$-Deep&0.81&0.87&0.84&0.68&0.73&0.71&0.69&0.75&0.72&0.57&0.63&0.60&0.53&0.55&0.54&0.49&0.43&0.46\\
\textbf{CellNuc-DETR}$_{base,4lvl}$&0.82&0.84&0.83&0.70&0.72&0.71&0.70&0.75&0.72&0.58&0.63&0.60&0.56&0.54&0.55&0.47&0.42&0.44\\
\textbf{CellNuc-DETR}$_{base,4lvl}$-Deep&0.81&0.86&0.83&0.69&0.73&0.71&0.69&0.75&0.72&0.57&0.62&0.59&0.54&0.55&0.55&0.47&0.43&0.45\\

\midrule

\textbf{CellNuc-DETR}$_{large,3lvl}$-Deep&0.82&0.85&0.83&0.70&0.73&0.72&0.72&0.76&0.74&0.57&0.63&0.60&0.56&0.56&0.56&0.51&0.41&0.45\\
\textbf{CellNuc-DETR}$_{large,4lvl}$-Deep&0.82&0.85&0.84&0.71&0.74&0.72&0.72&0.77&0.74&0.57&0.64&0.60&0.56&0.55&0.56&0.48&0.43&0.45\\

\bottomrule
\multicolumn{19}{l}{\small Other metrics are extracted from \cite{hörst2023cellvit}.}

\end{tabular}
\end{adjustbox}
\vskip -0.1in
\end{table}

The results in Table \ref{tab:pannuke} suggest that using the last three feature levels from the backbone is sufficiently accurate for both cell nuclei detection and classification. Omitting the first feature level improves the efficiency of the model by reducing the number of tokens processed by the deformable transformer. Specifically, the first feature level consists of $4\times4\text{px}$ image patches ($1 \mu \text{m}/\text{token}$ at an input resolution of $0.25 \mu m/\text{px}$). The second, third, and fourth levels correspond to $2 \mu \text{m}/\text{token}$, $4 \mu \text{m}/\text{token}$, and $8 \mu \text{m}/\text{token}$, respectively. Therefore, for a $256\times256\text{px}$ image, the first level contains 4096 tokens, the second has 1024, the third 256, and the fourth 64 tokens.

\begin{figure}
    \centering
    \includegraphics[width=\textwidth]{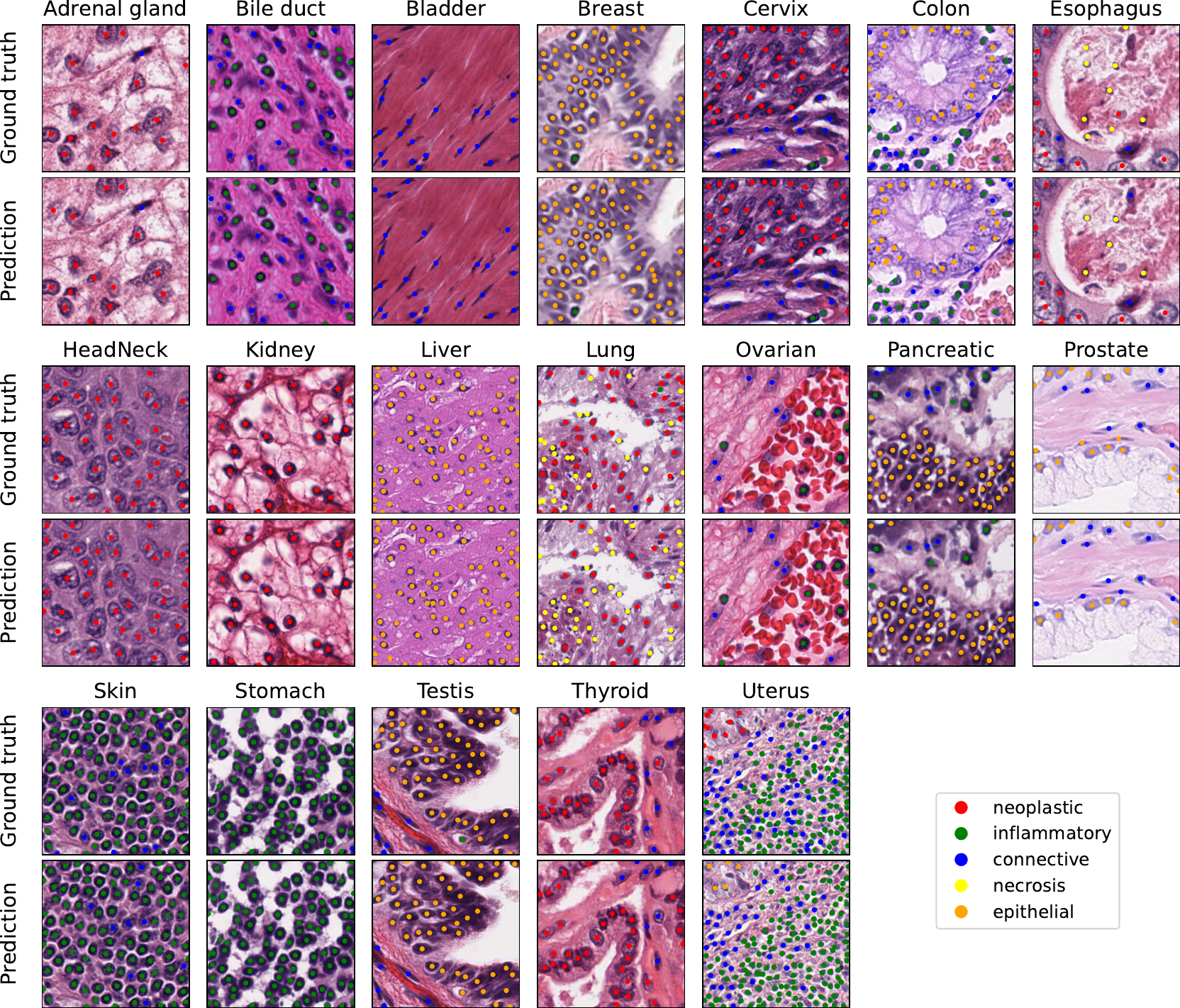}
    \caption{PanNuke ground truth and CellNuc-DETR$_{tiny,3lvl}$ predictions across tissues.}
    \label{fig:results:pannuke}
    \footnotesize{Visualization of CellNuc-DETR$_{tiny,3lvl}$ predictions for different test images of the PanNuke dataset. The images have been selected to showcase the differences between tissues, cell class distribution, cell density and stain variations.}
\end{figure}

By ignoring the first level, the number of tokens processed by the deformable transformer is reduced from 5440 to 1340. However, this reduction only decreases the computational load of the deformable transformer and does not impact the computational cost of the backbone. To address this, we explored training the model at a lower input resolution of $0.50 \mu m/\text{px}$, where cell nuclei are still easily identifiable. This change reduces the overall computational cost, as the input images are smaller. By using all four feature levels at $0.50 \mu m/\text{px}$, we ensure that the deformable transformer continues to receive feature maps at relevant scales ($2 \mu \text{m}/\text{token}$, $4 \mu \text{m}/\text{token}$, $8 \mu \text{m}/\text{token}$) while significantly reducing the backbone's computational burden.

Table \ref{tab:pannuke05} shows the results for CellNuc-DETR and CellViT on the PanNuke dataset, with training and evaluation conducted at a resolution of $0.50 \mu m/\text{px}$, averaged across the three folds. The model uses the Swin-tiny (\textit{tiny}) backbone and takes the four feature levels (\textit{4lvl}) to ensure the relevant information is taken. We experiment with both the larger model taking 900 decoder queries and six layers (Deep) and the smaller with 600 queries and three layers. The results demonstrate that CellNuc-DETR is highly effective, achieving significantly superior performance in both cell nuclei detection and classification tasks. Indeed, CellNuc-DETR shows only minimal performance degradation compared to the full-resolution models presented in Table \ref{tab:pannuke}, validating our previous observations.

The model maintains acceptable performance even in the most challenging classes, such as necrosis, which involve very small and underrepresented nuclei. These results suggest that using a lower resolution like $0.50 \mu m/\text{px}$ can be beneficial for speeding up inference time and reducing computational load in clinical scenarios while maintaining high accuracy.

\begin{table}[ht]
\vskip -0.1in
\centering
\caption{Detection and classification metrics on PanNuke dataset.}
\label{tab:pannuke05}
\vskip 0.1in
\begin{adjustbox}{width=\textwidth}
\begin{tabular}{l|ccc|ccc|ccc|ccc|ccc|cccccc}
\toprule
\multirow{2}{*}{\textbf{Method}} & \multicolumn{3}{c|}{\textbf{Detection}} & \multicolumn{3}{c|}{\textbf{Neoplastic}} & \multicolumn{3}{c|}{\textbf{Epithelial}} & \multicolumn{3}{c|}{\textbf{Inflammatory}} & \multicolumn{3}{c|}{\textbf{Connective}} & \multicolumn{3}{c}{\textbf{Necrosis}} \\
& $P_{det}$ & $R_{det}$ & $F_{det}$ & $P_{neo}$ & $R_{neo}$ & $F_{neo}$ & $P_{epi}$ & $R_{epi}$ & $F_{epi}$ & $P_{inf}$ & $R_{inf}$ & $F_{inf}$ & $P_{con}$ & $R_{con}$ & $F_{con}$ & $P_{nec}$ & $R_{nec}$ & $F_{nec}$ \\
\midrule

\textbf{CellViT$_{256}$}(0.50$\mu$m/px) \cite{hörst2023cellvit} & 0.86 & 0.60 & 0.71 & 0.72 & 0.59 & 0.65 & 0.71 & 0.58 & 0.64 & 0.60 & 0.38 & 0.47 & 0.53 & 0.32 & 0.40 & 0.43 & 0.04 & 0.07 \\
\textbf{CellViT$_{SAM-H}$}(0.50$\mu$m/px) \cite{hörst2023cellvit} & 0.88 & 0.63 & 0.73 & 0.74 & 0.62 & 0.67 & 0.74 & 0.61 & 0.67 & 0.60 & 0.42 & 0.49 & 0.56 & 0.34 & 0.42 & 0.49 & 0.04 & 0.08 \\

\midrule

\textbf{CellNuc-DETR}$_{tiny,4lvl}$(0.50$\mu$m/px)&0.79&0.86&0.82&0.66&0.73&0.70&0.67&0.76&0.72&0.56&0.61&0.58&0.53&0.54&0.54&0.42&0.42&0.42\\
\textbf{CellNuc-DETR}$_{tiny,4lvl}$-Deep(0.50$\mu$m/px)&0.80&0.86&0.83&0.67&0.73&0.70&0.67&0.76&0.71&0.56&0.62&0.59&0.54&0.54&0.54&0.46&0.40&0.42\\



\bottomrule
\multicolumn{19}{l}{\small Other metrics are extracted from \cite{hörst2023cellvit}.}

\end{tabular}
\end{adjustbox}
\vskip -0.1in
\end{table}

\begin{figure}
    \centering
    \includegraphics[width=0.8\textwidth]{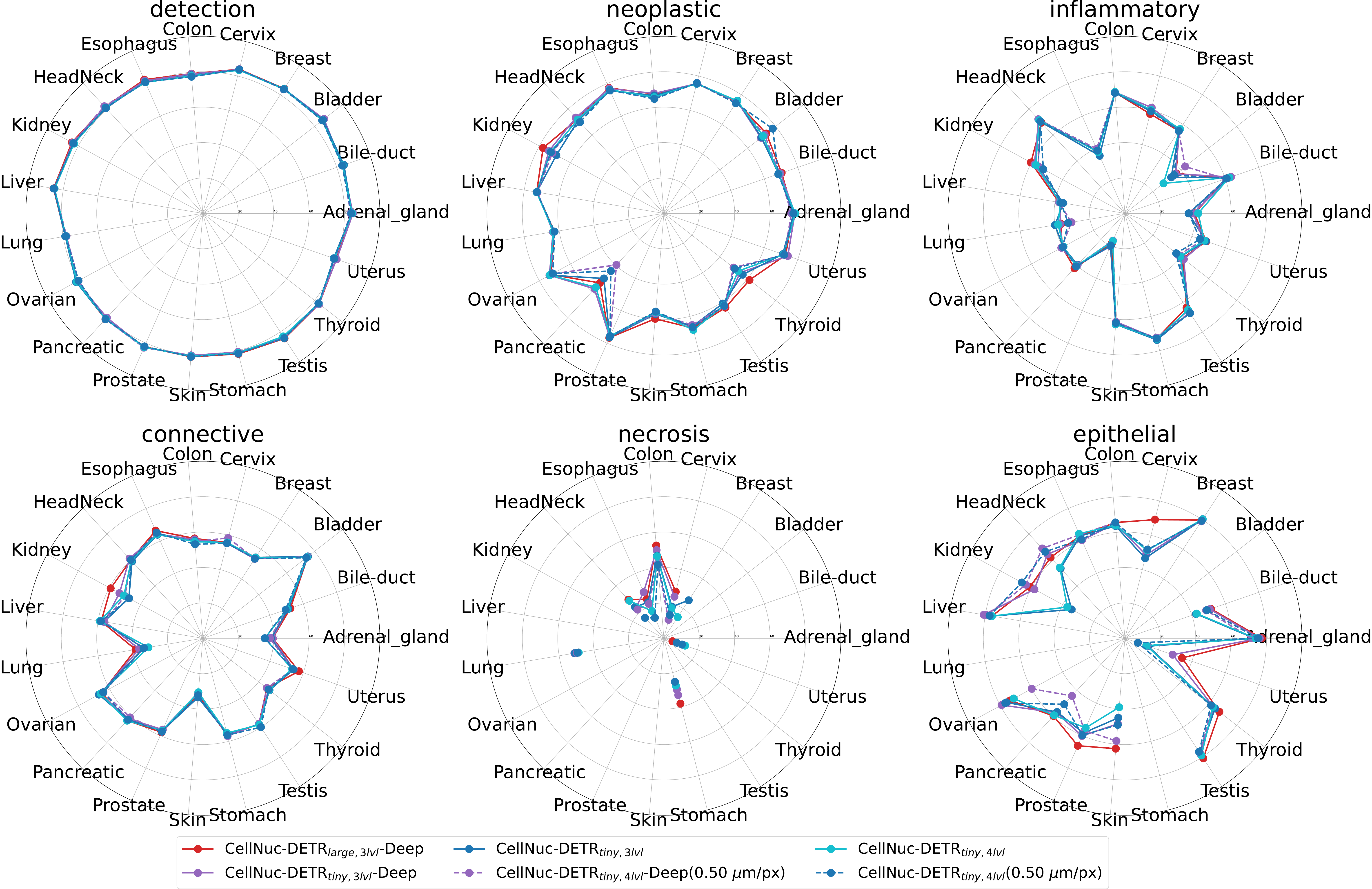}
    \caption{F1-Score for cell nuclei detection and classification across tissues.}
    \label{fig:results:radar}
\end{figure}

To better understand the performance of CellNuc-DETR, we conducted a detailed per-tissue analysis of cell nuclei detection and classification tasks. Figure \ref{fig:results:radar} illustrates the performance of CellNuc-DETR across the 19 tissues of PanNuke. Each line corresponds to the performance each CellNuc-DETR variation explored averaged across the three PanNuke folds. While similar performance is achieved for all model variations and scales, some tissues pose greater challenges than others.

cell nuclei detection performance is consistent across all tissues; however, classification performance varies significantly. For instance, epithelial cell classification shows poor results in tissues such as lung, kidney, cervix, bile duct, uterus, stomach, and skin. Inflammation classification also displays considerable variance, with F1-Scores above 60\% in tissues like kidney, head and neck, colon, bile duct, testis, stomach, and skin, but dropping to around 40\% for other tissues. Similarly, the classification of connective tissue is suboptimal in tissues such as lung and skin.

These discrepancies are likely due to variations in cell nucleus appearance across different tissues, as well as class imbalance within each tissue for each cell type. Tissues with more diverse or less abundant cell types may present significant challenges for accurate classification, contributing to the observed variability in model performance.

\subsubsection{Cross-dataset Evaluation}
\label{sec:results:numerical:cross}

To further validate the robustness of CellNuc-DETR, we performed a cross-dataset evaluation using CellNuc-DETR models trained on 80\% of the PanNuke dataset. These models were evaluated on the CoNSeP and MoNuSeg datasets without any additional fine-tuning. Given the differences in class definitions between PanNuke and CoNSeP, we mapped the classes as follows: neoplastic and epithelial cells from PanNuke were mapped to the epithelial class in CoNSeP, inflammatory cells were mapped directly as the same, connective tissue cells were mapped to the spindle-shaped class, and necrotic cells were mapped to the miscellaneous class.

The results in Table \ref{tab:consep} show that CellNuc-DETR significantly outperforms other state-of-the-art methods. Notable, these methods have been specifically trained on CoNSeP, whereas we are performing a cross-dataset evaluation for CellNuc-DETR.  This highlights the robustness of CellNuc-DETR, which is crucial for practical applications in digital pathology. The results obtained on the CoNSeP dataset are similar to the CellNuc-DETR performance on the colon tissue images of the PanNuke dataset reported in Figure \ref{fig:results:radar}. Additionally, we included results for models trained at a resolution of $0.50 \mu m/\text{px}$. Once again, the performance is very similar to the full-resolution models, further validating CellNuc-DETR's robustness across different datasets and scales and demonstrating its wide applicability. Interestingly, in the cross-domain evaluation, the size of the deformable transformer has a more significant impact on generalization performance, suggesting that model complexity plays a more critical role when dealing with cross-domain scenarios.



\begin{table}[ht]
\vskip -0.1in
\centering
\caption{Detection and classification F-Score on CoNSeP and MoNuSeg datasets. }
\label{tab:consep}
\vskip 0.1in
\begin{adjustbox}{width=0.9\textwidth}
\begin{tabular}{l|ccccc|c}
\toprule
\textbf{} & \multicolumn{5}{c|}{\textbf{CoNSeP}} & \textbf{MoNuSeg} \\
\cmidrule(lr){2-6} \cmidrule(lr){7-7}
\textbf{Method} & \textbf{Detection} & \textbf{Epithelial} & \textbf{Inflammatory} & \textbf{Spindle-shaped} & \textbf{Miscellaneous} & \textbf{Detection} \\
\midrule
\textbf{DIST*} \cite{naylor2018segmentation} & 0.71 & 0.62 & 0.53 & 0.51 & 0.00 & - \\
\textbf{Micro-Net*} \cite{raza2019micro} & 0.74 & 0.62 & 0.59 & 0.53 & 0.12 & - \\
\textbf{Mask-RCNN*} \cite{he2017mask} & 0.69 & 0.60 & 0.59 & 0.52 & 0.10 & - \\
\textbf{HoVer-Net*} \cite{graham2019hover} & 0.75 & 0.64 & 0.63 & 0.57 & 0.43 & - \\
\textbf{ACFormer*} \cite{Huang_2023_ICCV} & 0.74 & 0.64 & 0.64 & - & - & - \\
\midrule
\textbf{CellNuc-DETR}$_{tiny,3lvl}$ & 0.77 & 0.71 & 0.70 & 0.62 & 0.59 & 0.87 \\
\textbf{CellNuc-DETR}$_{tiny,3lvl}$-Deep & 0.78 & 0.72 & 0.72 & 0.63 & 0.63 & 0.88 \\
\textbf{CellNuc-DETR}$_{tiny,4lvl}$(0.50$\mu$m/px) & 0.75 & 0.69 & 0.71 & 0.60 & 0.52 & 0.87 \\
\textbf{CellNuc-DETR}$_{tiny,4lvl}$-Deep(0.50$\mu$m/px) & 0.76 & 0.69 & 0.73 & 0.61 & 0.59 & 0.87 \\


\bottomrule

\multicolumn{7}{l}{\small *Models trained on CoNSeP.}

\end{tabular}
\end{adjustbox}
\vskip -0.15in
\end{table}

\begin{figure}
    \centering
    \includegraphics[width=\textwidth]{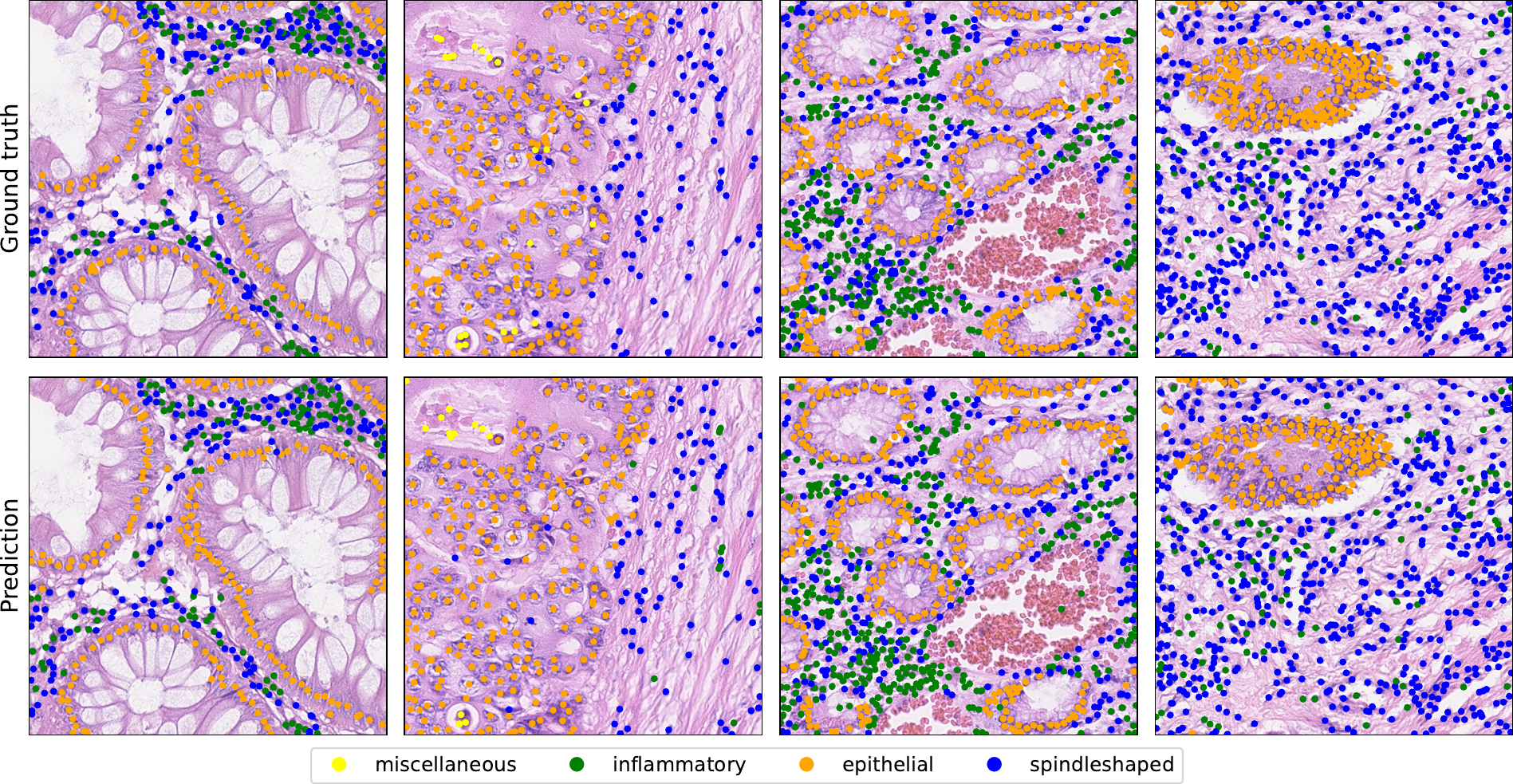}
    \caption{CoNSeP ground truth and CellNuc-DETR$_{tiny,3lvl}$ predictions.}
    \label{fig:results:consep}
\end{figure}

\subsection{Inference Time}
\label{sec:results:time}

In this section, we evaluate the inference time of the CellNuc-DETR pipeline on the WSIs obtained from TCGA. We compare the performance of CellNuc-DETR with other state-of-the-art methods, such as HoVer-NeXt that reported significantly higher speeds than HoVer-Net and CellViT, to assess both speed and efficiency in processing large-scale histopathological data. Additionally, we explore the influence of various design components within the CellNuc-DETR pipeline on its inference performance, providing insights into how different configurations can impact overall efficiency.

\subsubsection{Pipeline Comparison}
\label{sec:results:time:sota}

\begin{figure}[h]
    \centering
    \includegraphics[width=\textwidth]{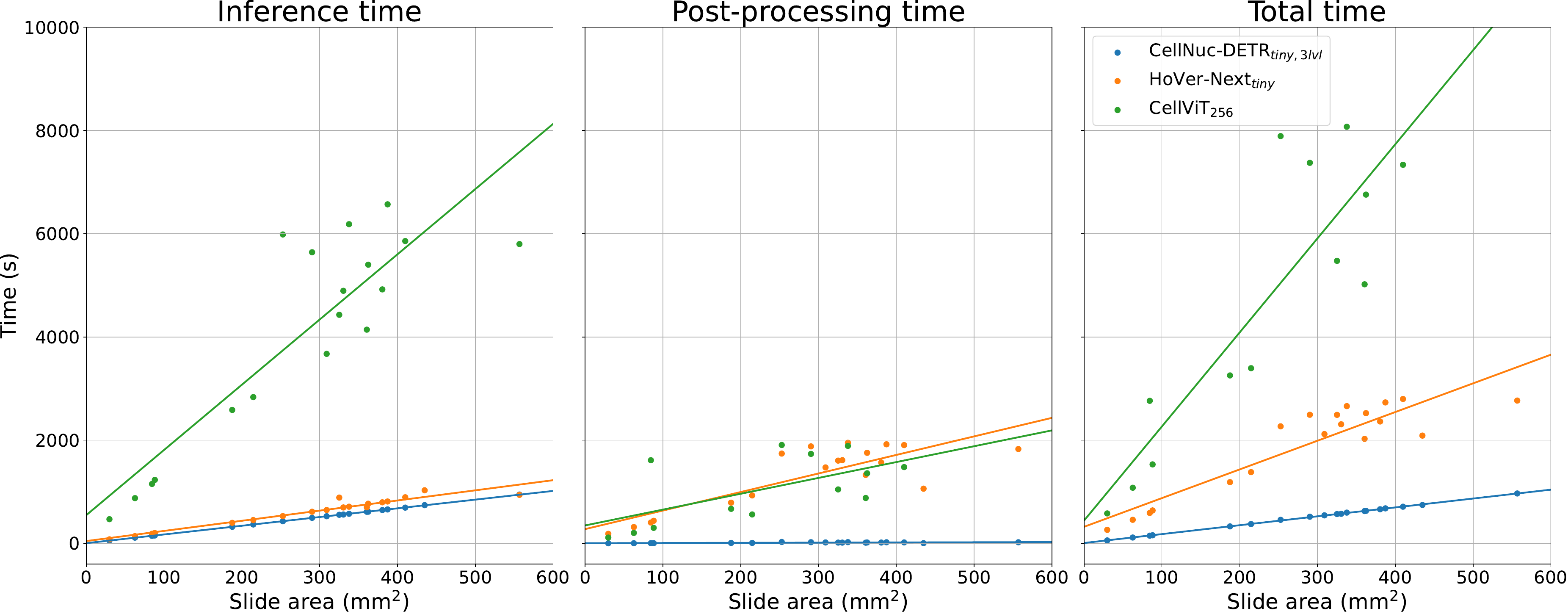}
    \caption{Pipeline inference, post-processing and total time as function of the slide area.}
    \label{fig:results:inference-sota}
    \footnotesize{Pipeline inference time refers to the model processing of the individual tiles extracted from the WSI, whereas the post-processing time is the time spent combining the predictions and returning the final results.}
\end{figure}

In Figure \ref{fig:results:inference-sota}, we compare the inference and post-processing times on WSIs for three different pipelines: CellNuc-DETR, Cell-ViT, and HoVer-NeXt. The three pipelines are run with the 20 WSIs extracted from TCGA and the times are reported as function of the slides' area. Each dot in the plot corresponds to a slide, we have also included the regression lines for a more comprehensive visualization.

We select models trained on PanNuke that operate at the same magnification of 0.25 $\mu$m/px. For Cell-ViT, we chose the smaller Cell-ViT-256 model, as it is faster than Cell-ViT-SAM-H, and conducted inference on tiles of $1024 \times 1024$ pixels, as suggested in the original paper. For HoVer-NeXt, we used the configuration recommended in the paper for obtaining the metrics in Table \ref{tab:pannuke}: ConvNeXtv2-tiny backbone with four test-time augmentations. Finally, for CellNuc-DETR, we used a Swin-tiny backbone, which has a similar computational complexity to ConvNeXtv2-tiny, and selected the model with 3 backbone levels and 3 layers.

We evaluate the inference time, which corresponds to the time spent by the model in processing all individual tiles, as well as the post-processing time required to combine predictions and obtain the final results. The pipelines have been run on a NVIDIA A100 24GB GPU, within a cluster equipped with 8 CPU cores and 128GB RAM.

CellNuc-DETR is the fastest method in both inference time and, especially, post-processing time. As observed, CellNuc-DETR and HoVer-NeXt are significantly faster than Cell-ViT in processing tiles, with CellNuc-DETR being slightly faster. However, HoVer-NeXt's speed comes at the expense of lower accuracy, whereas CellNuc-DETR achieves state-of-the-art performance, as reported in Table \ref{tab:pannuke}. Notably, the post-processing time for CellNuc-DETR is forty times lower than that of HoVer-NeXt, demonstrating a substantial efficiency advantage.

\subsubsection{Speeding Up CellNuc-DETR}
\label{sec:results:time:scale}

\begin{figure}
    \centering
    \includegraphics[width=0.6\textwidth]{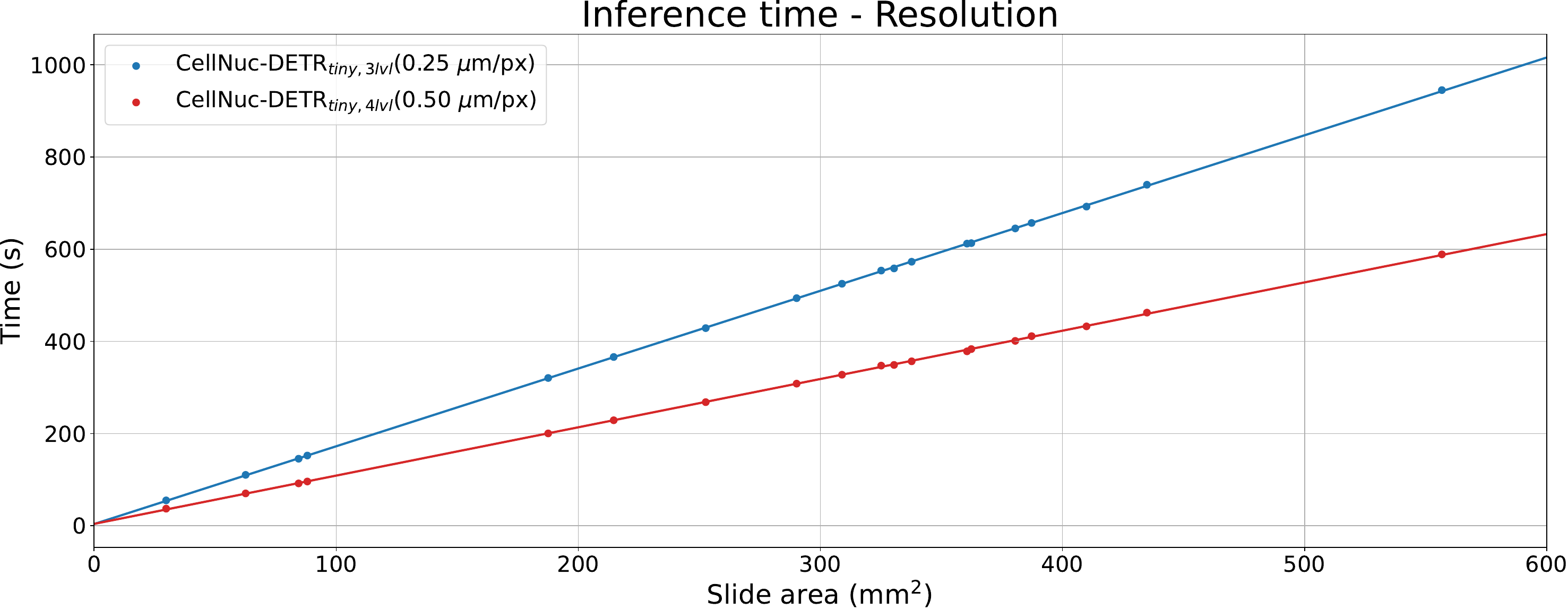}
    \caption{Effect of the input image resolution in the pipeline inference time.}
    \label{fig:results:inference-scale}
\end{figure}

One effective strategy to accelerate the inference time of CellNuc-DETR is to operate at a lower resolution, specifically $0.50 \mu m/\text{px}$, compared to the original $0.25 \mu m/\text{px}$. As discussed in Section \ref{sec:results:numerical:pannuke}, reducing the input image resolution directly lowers the overall computational complexity of the model. This reduction is achieved without compromising the model’s ability to capture essential features, as by leveraging the four backbone levels, the deformable transformer still processes the relevant resolutions necessary for accurate cell nuclei detection and classification. Indeed, the results in Table \ref{tab:pannuke05} show a minimal accuracy drop when working at $0.50 \mu m/\text{px}$.

In this section, we present the inference times of CellNuc-DETR when running at $0.50 \mu m/\text{px}$, as shown in Figure \ref{fig:results:inference-scale}. The results are remarkable, with the inference time reduced by a factor of 2 compared to the $0.25 \mu m/\text{px}$ configuration. This significant reduction highlights the practical benefits of using lower resolutions for faster processing, especially in clinical settings where time efficiency is critical. The ability to maintain high accuracy while drastically improving speed underscores the effectiveness of this strategy and reinforces the adaptability of CellNuc-DETR for various real-world applications.

\begin{figure}[h]
    \centering
    \includegraphics[width=0.6\textwidth]{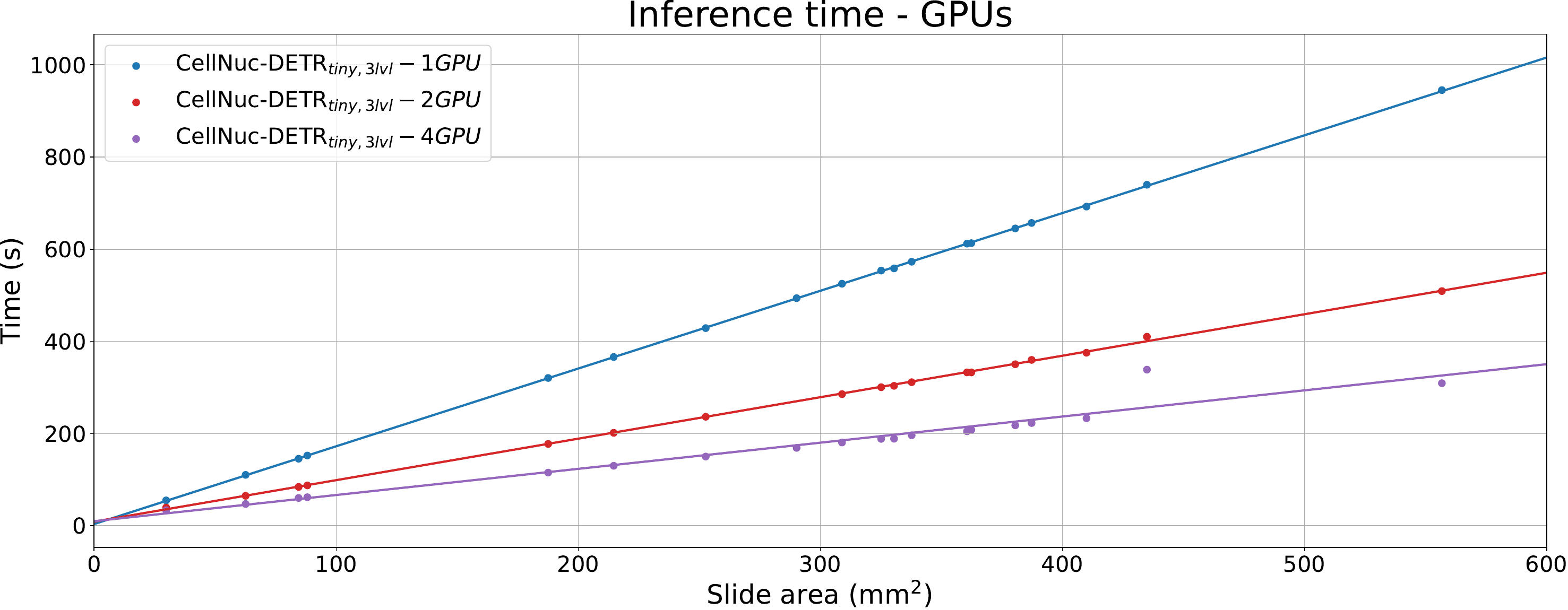}
    \caption{Effect of the number of GPUs in the pipeline inference time.}
    \label{fig:results:inference-gpu}
\end{figure}

Additionally, we have devised our pipeline to run in a distributed mode across multiple GPUs. This approach distributes the workload evenly across the GPUs, enabling parallel processing of tiles within WSIs. By leveraging multiple GPUs, the inference time is significantly reduced, making the pipeline more scalable and suitable for large-scale clinical applications or research environments requiring high throughput. The results of this distributed processing (at 0.25$\mu$m/px) are shown in Figure \ref{fig:results:inference-gpu}, demonstrating further improvements in inference efficiency compared to single GPU setups.

\subsubsection{Model's Size Effect}
\label{sec:results:time:size}

Finally, in this section we explore how various components of the CellNuc-DETR model affect inference time, specifically examining the impact of the backbone choice, the number of feature levels extracted from the backbone, as well as the depth and number of queries of the deformable transformer. As expected, the more sophisticated the model configuration, the slower the inference time. 

\begin{figure}
    \centering
    \includegraphics[width=0.6\textwidth]{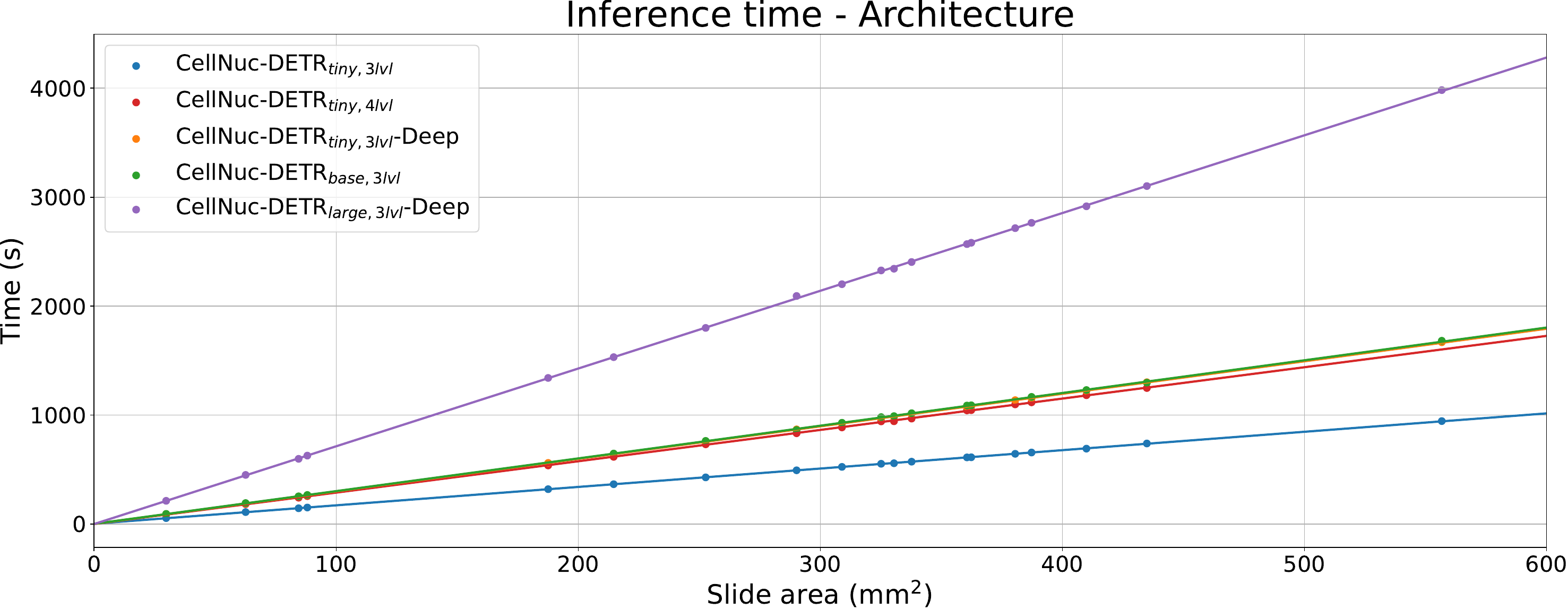}
    \caption{Effect of the model components in the pipeline inference time.}
    \label{fig:results:inference-model}
\end{figure}

As shown in Figure \ref{fig:results:inference-model}, including all four feature levels from the backbone, which increases the number of tokens fed into the transformer, has a similar effect on slowing down the model as increasing the depth of the transformer itself. Among the components analyzed, the choice of backbone has the most significant impact on inference time. For example, although Table \ref{tab:pannuke} shows that using a Swin-Large backbone results in improved classification performance, this improvement comes at the cost of much higher inference times. This illustrates the trade-off between model complexity and computational efficiency, highlighting the importance of carefully selecting model components based on the specific requirements of the application, whether that be higher accuracy or faster processing.

\section{Discussion}
\label{sec:discussion}

\paragraph{Detection vs Segmentation} In this paper, we propose a shift in the paradigm for extracting cell information from WSIs, moving from traditional cell segmentation approaches to cell nuclei detection. The motivation for this shift is that the primary tasks in digital pathology are cell nuclei detection and classification, where the focus is on identifying and labeling cells rather than generating pixel-wise segmentations. This shift is critical because detection methods are inherently faster and more efficient than segmentation methods, which require more computational resources, especially during post-processing.

We perform a detailed comparison between CellNuc-DETR and two segmentation methods optimized for cell nuclei detection and classification: CellViT, which focuses on achieving high detection and classification performance, and HoVer-NeXt, which is optimized for faster inference. In Table \ref{tab:pannuke}, we present the performance of CellNuc-DETR, demonstrating that it outperforms both methods in terms of cell classification and performs comparably to CellViT for cell nuclei detection. HoVer-NeXt, on the other hand, shows slightly lower performance in both tasks. Although CellViT provides a boost in performance compared to HoVer-NeXt, this improvement comes at the expense of a significantly slower inference pipeline, as illustrated in Figure \ref{fig:results:inference-sota}.

In contrast, the CellNuc-DETR pipeline achieves significantly faster inference times than both HoVer-NeXt and CellViT due to the efficiency of the detection task. Moreover, we have further optimized the pipeline, as discussed in Section \ref{sec:results:time}, by exploring different design components and configurations. As a result, we have developed a cell nuclei detection pipeline that is not only more accurate but also faster than the current state-of-the-art segmentation methods.

\begin{figure}
    \centering
    \includegraphics[width=0.6\textwidth]{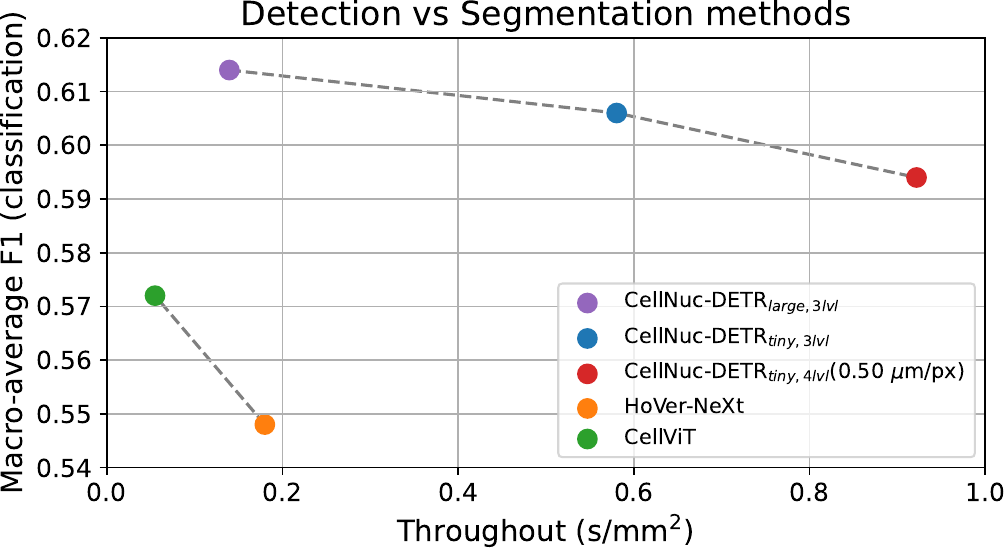}
    \caption{Throughout and performance of detection and segmentation pipelines.}
    \label{fig:discussion:detection-vs-segmentation}
\end{figure}

These advantages are clearly visualized in Figure \ref{fig:discussion:detection-vs-segmentation}, where we compare the macro-average F1 score for cell classification against the throughput (mm$^2$/s) of the pipelines. The figure shows that detection-based methods (CellNuc-DETR$_{tiny,3lvl}$, CellNuc-DETR$_{large,3lvl}$ and CellNuc-DETR$_{tiny,4lvl}$(0.50$\mu$m/px)) outperform segmentation-based methods (CellViT and HoVer-NeXt) both in accuracy and efficiency, establishing a new standard for cell nuclei detection and classification in digital pathology.

\paragraph{In-device sliding window} One of the components of CellNuc-DETR inference pipeline on WSIs is the use of an in-device overlapping window approach. This method involves dividing large image tiles into smaller, overlapping windows and processing them in parallel directly on the device. Firstly, retrieving large image tiles and processing them in-device with a sliding window rather than directly retrieving small patches reduces the number of I/O interactions with the slide, thereby decreasing data loading times. This minimizes the latency associated with accessing data from disk, which can be bottleneck in processing WSIs. Additionally, performing the sliding window operation in-device slightly reduces data transfer times compared to applying the sliding window on the CPU before sending data to the GPU. The overlap between windows results in redundant data when processed externally, increasing the amount of information transferred to the GPU. By conducting the sliding window operation directly on the device, CellNuc-DETR avoids this redundancy. Overall, the in-device sliding window approach enhances both speed and resource efficiency, making it a key component of the CellNuc-DETR pipeline for effective WSI analysis in digital pathology.

\paragraph{Dealing with edge cells}  During inference, the challenge of handling edge cells arises both within windows inside a tile and between tiles across a slide. Our approach leverages the overlapping nature of both windows and tiles to address this issue using a consistent strategy: we only retain cells whose centroids fall within the central crop of the image. The width of the margins is set to half of the overlap, ensuring that any centroid detected outside these borders is discarded, as it will be captured in the overlapping region of the adjacent tile. By setting the overlap between windows equal to the overlap between tiles, we ensure that the results remain agnostic to tile size. By default, we set both the window and tile overlaps to 64 pixels.

This approach is feasible because we are focused solely on the centroid and labels of the cell nuclei, rather than their shape. This process is significantly simpler and more efficient than the complex post-processing required by segmentation methods, highlighting one of the advantages of detection over segmentation.

While this method is an approximation, it effectively addresses the issue in most cases. A potential limitation occurs when a nucleus is larger than the overlap in the direction perpendicular to the tile partition and fully occupies the overlap between tiles, which could lead to duplicated detections. However, such instances are rare; only 0.3\% of the nuclei in the PanNuke dataset exceed 64 pixels in size. Indeed, the effectiveness of this window and tile merging strategy is empirically demonstrated by the results on the CoNSeP dataset, further validating our approach.

\section{Conclusions}
We proposed CellNuc-DETR, a detection-based approach for cell analysis in histopathological WSIs, shifting from the traditional segmentation methods used in digital pathology. Our method focuses on directly detecting and classifying cell nuclei, avoiding the computational burden of segmentation. Experiments on the PanNuke dataset demonstrate that CellNuc-DETR achieves state-of-the-art performance in both cell nuclei detection and classification. Cross-dataset evaluations on CoNSeP and MoNuSeg further confirm its robustness and generalization ability across diverse datasets and conditions.

In addition to its superior accuracy, CellNuc-DETR significantly enhances inference efficiency on large WSIs, being twice as fast as the fastest segmentation-based method, HoVer-NeXt, while achieving better accuracy. Compared to CellViT, CellNuc-DETR provides higher classification accuracy and operates approximately ten times more efficiently during WSI inference. These advantages make CellNuc-DETR highly suitable for clinical and high-throughput research settings where both speed and accuracy are essential.

Overall, CellNuc-DETR offers a new standard for cell analysis in digital pathology by combining high performance with computational efficiency. Future work could explore further optimizations for clinical applications and extend the method to additional histopathological tasks.

\bibliography{refs}

\begin{thebibliography}{10}

\bibitem{lewis2023automated}
Joshua~E Lewis, Conrad~W Shebelut, Bradley~R Drumheller, Xuebao Zhang, Nithya Shanmugam, Michel Attieh, Michael~C Horwath, Anurag Khanna, Geoffrey~H Smith, David~A Gutman, et~al.
\newblock An automated pipeline for differential cell counts on whole-slide bone marrow aspirate smears.
\newblock {\em Modern Pathology}, 36(2):100003, 2023.

\bibitem{lara2021quantitative}
Haydee Lara, Zaibo Li, Esther Abels, Famke Aeffner, Marilyn~M Bui, Ehab~A ElGabry, Cleopatra Kozlowski, Michael~C Montalto, Anil~V Parwani, Mark~D Zarella, et~al.
\newblock Quantitative image analysis for tissue biomarker use: a white paper from the digital pathology association.
\newblock {\em Applied Immunohistochemistry \& Molecular Morphology}, 29(7):479--493, 2021.

\bibitem{sobhani2021artificial}
Faranak Sobhani, Ruth Robinson, Azam Hamidinekoo, Ioannis Roxanis, Navita Somaiah, and Yinyin Yuan.
\newblock Artificial intelligence and digital pathology: Opportunities and implications for immuno-oncology.
\newblock {\em Biochimica et Biophysica Acta (BBA)-Reviews on Cancer}, 1875(2):188520, 2021.

\bibitem{pati2022hierarchical}
Pushpak Pati, Guillaume Jaume, Antonio Foncubierta-Rodriguez, Florinda Feroce, Anna~Maria Anniciello, Giosue Scognamiglio, Nadia Brancati, Maryse Fiche, Estelle Dubruc, Daniel Riccio, et~al.
\newblock Hierarchical graph representations in digital pathology.
\newblock {\em Medical image analysis}, 75:102264, 2022.

\bibitem{pina2022self}
Oscar Pina and Ver{\'o}nica Vilaplana.
\newblock Self-supervised graph representations of wsis.
\newblock In {\em Geometric Deep Learning in Medical Image Analysis}, pages 107--117. PMLR, 2022.

\bibitem{wang2024breast}
Kang Wang, Feiyang Zheng, Lan Cheng, Hong-Ning Dai, Qi~Dou, and Jing Qin.
\newblock Breast cancer classification from digital pathology images via connectivity-aware graph transformer.
\newblock {\em IEEE Transactions on Medical Imaging}, 2024.

\bibitem{garcia2016trying}
Marcial Garc{\'\i}a-Rojo and Jaume Ordi.
\newblock Trying to understand digital pathology before we move to computational pathology.
\newblock {\em Pathobiology}, 83(2-3):57--60, 2016.

\bibitem{graham2019hover}
Simon Graham, Quoc~Dang Vu, Shan E~Ahmed Raza, Ayesha Azam, Yee~Wah Tsang, Jin~Tae Kwak, and Nasir Rajpoot.
\newblock Hover-net: Simultaneous segmentation and classification of nuclei in multi-tissue histology images.
\newblock {\em Medical Image Analysis}, page 101563, 2019.

\bibitem{hörst2023cellvit}
Fabian Hörst, Moritz Rempe, Lukas Heine, Constantin Seibold, Julius Keyl, Giulia Baldini, Selma Ugurel, Jens Siveke, Barbara Grünwald, Jan Egger, and Jens Kleesiek.
\newblock Cellvit: Vision transformers for precise cell segmentation and classification, 2023.

\bibitem{baumann2024hover}
Elias Baumann, Bastian Dislich, Josef~Lorenz Rumberger, Iris~D Nagtegaal, Maria~Rodriguez Martinez, and Inti Zlobec.
\newblock Hover-next: A fast nuclei segmentation and classification pipeline for next generation histopathology.
\newblock In {\em Medical Imaging with Deep Learning}, 2024.

\bibitem{pina2024cell}
Oscar Pina, Eduard Dorca, and Veronica Vilaplana.
\newblock Cell-detr: Efficient cell detection and classification in wsis with transformers.
\newblock In {\em Medical Imaging with Deep Learning}, 2024.

\bibitem{prangemeier2020c}
Tim Prangemeier, Christoph Reich, and Heinz Koeppl.
\newblock Attention-based transformers for instance segmentation of cells in microstructures.
\newblock 2020.

\bibitem{carion2020end}
Nicolas Carion, Francisco Massa, Gabriel Synnaeve, Nicolas Usunier, Alexander Kirillov, and Sergey Zagoruyko.
\newblock End-to-end object detection with transformers.
\newblock In {\em European conference on computer vision}, pages 213--229. Springer, 2020.

\bibitem{kumar2019multi}
Neeraj Kumar, Ruchika Verma, Deepak Anand, Yanning Zhou, Omer~Fahri Onder, Efstratios Tsougenis, Hao Chen, Pheng-Ann Heng, Jiahui Li, Zhiqiang Hu, et~al.
\newblock A multi-organ nucleus segmentation challenge.
\newblock {\em IEEE transactions on medical imaging}, 39(5):1380--1391, 2019.

\bibitem{dosovitskiy2020vit}
Alexey Dosovitskiy, Lucas Beyer, Alexander Kolesnikov, Dirk Weissenborn, Xiaohua Zhai, Thomas Unterthiner, Mostafa Dehghani, Matthias Minderer, Georg Heigold, Sylvain Gelly, Jakob Uszkoreit, and Neil Houlsby.
\newblock An image is worth 16x16 words: Transformers for image recognition at scale.
\newblock {\em ICLR}, 2021.

\bibitem{liu2021swin}
Ze~Liu, Yutong Lin, Yue Cao, Han Hu, Yixuan Wei, Zheng Zhang, Stephen Lin, and Baining Guo.
\newblock Swin transformer: Hierarchical vision transformer using shifted windows.
\newblock In {\em Proceedings of the IEEE/CVF International Conference on Computer Vision (ICCV)}, 2021.

\bibitem{kuhn1955hungarian}
Harold~W Kuhn.
\newblock The hungarian method for the assignment problem.
\newblock {\em Naval research logistics quarterly}, 2(1-2):83--97, 1955.

\bibitem{zhu2020deformable}
Xizhou Zhu, Weijie Su, Lewei Lu, Bin Li, Xiaogang Wang, and Jifeng Dai.
\newblock Deformable detr: Deformable transformers for end-to-end object detection.
\newblock {\em arXiv preprint arXiv:2010.04159}, 2020.

\bibitem{anglada2024enhancing}
David Anglada-Rotger, Julia Sala, Ferran Marques, Philippe Salembier, and Montse Pard{\`a}s.
\newblock Enhancing ki-67 cell segmentation with dual u-net models: A step towards uncertainty-informed active learning.
\newblock In {\em Proceedings of the IEEE/CVF Conference on Computer Vision and Pattern Recognition}, pages 5026--5035, 2024.

\bibitem{ronneberger2015u}
Olaf Ronneberger, Philipp Fischer, and Thomas Brox.
\newblock U-net: Convolutional networks for biomedical image segmentation.
\newblock In {\em Medical image computing and computer-assisted intervention--MICCAI 2015: 18th international conference, Munich, Germany, October 5-9, 2015, proceedings, part III 18}, pages 234--241. Springer, 2015.

\bibitem{gamper2020pannuke}
Jevgenij Gamper, Navid~Alemi Koohbanani, Ksenija Benes, Simon Graham, Mostafa Jahanifar, Syed~Ali Khurram, Ayesha Azam, Katherine Hewitt, and Nasir Rajpoot.
\newblock Pannuke dataset extension, insights and baselines.
\newblock {\em arXiv preprint arXiv:2003.10778}, 2020.

\bibitem{deng2009imagenet}
Jia Deng, Wei Dong, Richard Socher, Li-Jia Li, Kai Li, and Li~Fei-Fei.
\newblock Imagenet: A large-scale hierarchical image database.
\newblock In {\em 2009 IEEE conference on computer vision and pattern recognition}, pages 248--255. Ieee, 2009.

\bibitem{lin2014microsoft}
Tsung-Yi Lin, Michael Maire, Serge Belongie, James Hays, Pietro Perona, Deva Ramanan, Piotr Doll{\'a}r, and C~Lawrence Zitnick.
\newblock Microsoft coco: Common objects in context.
\newblock In {\em Computer Vision--ECCV 2014: 13th European Conference, Zurich, Switzerland, September 6-12, 2014, Proceedings, Part V 13}, pages 740--755. Springer, 2014.

\bibitem{10.1117/12.2293048}
David Tellez, Maschenka Balkenhol, Nico Karssemeijer, Geert Litjens, Jeroen van~der Laak, and Francesco Ciompi.
\newblock {H and E stain augmentation improves generalization of convolutional networks for histopathological mitosis detection}.
\newblock In John~E. Tomaszewski and Metin~N. Gurcan, editors, {\em Medical Imaging 2018: Digital Pathology}, volume 10581, page 105810Z. International Society for Optics and Photonics, SPIE, 2018.

\bibitem{ruifrok2001quantification}
Arnout~C Ruifrok, Dennis~A Johnston, et~al.
\newblock Quantification of histochemical staining by color deconvolution.
\newblock {\em Analytical and quantitative cytology and histology}, 23(4):291--299, 2001.

\bibitem{goode2013openslide}
Adam Goode, Benjamin Gilbert, Jan Harkes, Drazen Jukic, and Mahadev Satyanarayanan.
\newblock Openslide: A vendor-neutral software foundation for digital pathology.
\newblock {\em Journal of pathology informatics}, 4(1):27, 2013.

\bibitem{he2016deep}
Kaiming He, Xiangyu Zhang, Shaoqing Ren, and Jian Sun.
\newblock Deep residual learning for image recognition.
\newblock In {\em Proceedings of the IEEE conference on computer vision and pattern recognition}, pages 770--778, 2016.

\bibitem{naylor2018segmentation}
Peter Naylor, Marick La{\'e}, Fabien Reyal, and Thomas Walter.
\newblock Segmentation of nuclei in histopathology images by deep regression of the distance map.
\newblock {\em IEEE transactions on medical imaging}, 38(2):448--459, 2018.

\bibitem{he2017mask}
Kaiming He, Georgia Gkioxari, Piotr Doll{\'a}r, and Ross Girshick.
\newblock Mask r-cnn.
\newblock In {\em Proceedings of the IEEE international conference on computer vision}, pages 2961--2969, 2017.

\bibitem{raza2019micro}
Shan E~Ahmed Raza, Linda Cheung, Muhammad Shaban, Simon Graham, David Epstein, Stella Pelengaris, Michael Khan, and Nasir~M Rajpoot.
\newblock Micro-net: A unified model for segmentation of various objects in microscopy images.
\newblock {\em Medical image analysis}, 52:160--173, 2019.

\bibitem{Huang_2023_ICCV}
Junjia Huang, Haofeng Li, Xiang Wan, and Guanbin Li.
\newblock Affine-consistent transformer for multi-class cell nuclei detection.
\newblock In {\em Proceedings of the IEEE/CVF International Conference on Computer Vision (ICCV)}, pages 21384--21393, October 2023.

\end{thebibliography}
\bibliographystyle{unsrt}

\appendix

\section{Implementation Details}

This section outlines the specifics of the training and inference processes, including the computational resources utilized, hyperparameters, and data augmentation techniques. All models were implemented in PyTorch, based on the official repositories of Deformable-DETR \cite{zhu2020deformable} and DETR \cite{carion2020end}.

\subsection{Training and Evaluation Details}

The training was conducted on a cluster equipped with four NVIDIA Quadro 16GB GPUs, 16GB of RAM, and 8 CPUs. The training hyperparameters, such as learning rate, batch size, and other optimization parameters, are summarized in Table \ref{tab:hyperparams}. The batch size indicated in the table corresponds to the batch size per GPU. The learning rate was linearly scaled according to the batch size and the number of devices, based on the original batch size of 16 from the Deformable-DETR paper, to ensure consistent optimization across different hardware configurations.

Data augmentation techniques employed during training—such as rotation, flipping, and stain augmentations—are detailed in Table \ref{tab:data-augmentation}. These augmentations help enhance model generalization by simulating various staining protocols and slide preparation techniques.

The models for PanNuke dataset results presented in Tables \ref{tab:pannuke} and \ref{tab:pannuke05} were trained using the standard three-fold split provided with the dataset. Although we did not perform an exhaustive hyperparameter search, the default hyperparameters were used, with validation sets from each fold employed to avoid overfitting and to determine the optimal confidence threshold for filtering out no-object queries. The threshold was selected by maximizing the harmonic mean between the detection F1-score and the macro-average classification F1-score, ensuring balanced performance across both metrics.

For cross-dataset evaluations in Table \ref{tab:consep}, models were trained on 80\% of the PanNuke dataset, with 80\% taken from each fold to create an expanded training set. The remaining 20\% served as a validation set to prevent overfitting and to fine-tune the confidence threshold. For the MoNuSeg dataset, which includes only detection labels, the threshold was chosen to maximize the detection F1-score on the PanNuke validation set (20\%). Although using CoNSeP and MoNuSeg training sets could potentially yield better thresholds, we opted to base this decision solely on the PanNuke dataset to maintain a strict cross-dataset evaluation protocol.

\subsection{Inference Details}

Inference on WSIs for CellNuc-DETR, CellViT, and HoVerNeXt was conducted using an NVIDIA GeForce 24GB GPU, along with 128GB of RAM and 8 CPUs. Additional hyperparameters related to the inference process, including tile sizes, overlaps, and batch processing strategies, are provided in Table Z. These parameters were carefully selected to strike a balance between computational efficiency and model accuracy.

\begin{table}[h]
\centering
\caption{Training hyperparameters}
\label{tab:hyperparams}
\begin{tabular}{l|cc}
\toprule
\textbf{Group} & \textbf{Hyperparameter} & \textbf{Value} \\
\midrule
\multirow{4}{*}{\textbf{Solver}} & Epochs & 100 \\
 & Base LR & 2e-4 \\
 & Batch Size & 2 \\
 & LR Drop & 0.10 \\
 & LR Steps & 70, 90 \\
\midrule
\multirow{3}{*}{\textbf{Matcher}} & $\lambda_{\text{giou}}$ & 2 \\
 & $\lambda_{\text{bbox}}$ & 2 \\
 & $\lambda_{\text{focal}}$ & 5 \\
\midrule
\multirow{4}{*}{\textbf{Loss}} & $\lambda_{\text{giou}}$ & 2 \\
 & $\lambda_{\text{bbox}}$ & 1 \\
 & $\lambda_{\text{focal}}$ & 5 \\
 & $\alpha_{\text{focal}}$ & 0.25 \\
\bottomrule

\end{tabular}
\end{table}

\begin{table}[h]
\centering
\caption{Data augmentation hyperparameters}
\label{tab:data-augmentation}
\begin{tabular}{lc|cc}
\toprule
\textbf{Augmentation} & \textbf{Probability} & \textbf{Hyperparameter} & \textbf{Value} \\ 
\midrule
\multirow{2}{*}{Elastic} & \multirow{2}{*}{0.2} & $\alpha$ & 0.5 \\ 
 &  & $\sigma$ & 0.25 \\ 
\midrule
Horizontal Flip & 0.5 & - & - \\ 
\midrule
Vertical Flip & 0.5 & - & - \\ 
\midrule
\multirow{1}{*}{Rotate} & \multirow{1}{*}{1.0} & angles & [0,90,180,270] \\ 
\midrule
\multirow{2}{*}{Blur} & \multirow{2}{*}{0.2} & kernel size & 9 \\ 
 &  & $\sigma$ & [0.2, 1.0] \\ 
\midrule
\multirow{2}{*}{HED Transform} & \multirow{2}{*}{0.2} & $\alpha$ & 0.04 \\ 
 &  & $\beta$ & 0.04 \\ 
\midrule
\multirow{2}{*}{Resized Crop} & \multirow{2}{*}{0.2} & size & 256 \\ 
 &  & scale & [0.8, 1.0] \\ 
\bottomrule
\end{tabular}
\end{table}

\end{document}